\documentclass[journal]{IEEEtran}
\ifCLASSINFOpdf
\else
\fi
\usepackage{amssymb}
\usepackage{amsmath}
\usepackage{algorithm}
\usepackage{algorithmic}
\usepackage{hyperref}
\usepackage{tikz}
 
\usepackage{graphicx}
\usepackage{booktabs}
\usepackage{xspace}

\usepackage{multirow}
\usepackage{makecell}
\DeclareMathOperator*{\argmin}{arg\,min}
\usepackage{ mathrsfs }
\usepackage{graphicx, adjustbox}
\usepackage{array}
\usepackage{setspace}
\usepackage{enumitem}
\usepackage{breqn}
\usepackage{bm}
\usepackage{ dsfont }
\usepackage{ upgreek }
\usepackage{textcomp}
\usepackage{multirow}
\usepackage{algorithm}
\usepackage{algorithmic}
\usepackage{caption}
\usepackage{pgfplots}
\usepackage{tkz-euclide}
\usepackage{subcaption}
\usepackage{tikz}{\tiny }
\usepackage{balance}
\usepackage{ tipa }

\usepackage{fancyhdr}
\usepackage{ upgreek }
\usepackage{soul}
\usepackage{breqn}
\usepackage[english]{babel}

\addto\captionsenglish{}
\allowdisplaybreaks

\usepackage{amsthm}
\newtheorem{theorem}{Theorem}
\newtheorem{lemma}{Lemma}

\newtheorem{remark}{Remark}

\newtheorem{assumption}{Assumption}

\usepackage{fancyhdr}
\usepackage{ upgreek }
\usepackage{soul}
\makeatletter
\newcommand{\vast}{\bBigg@{4}}

\newcommand{\Vast}{\bBigg@{5}}
\begin{document}

\title{Probabilistic Federated  Learning on Uncertain and Heterogeneous Data with Model Personalization}

\author{Ratun Rahman
        and Dinh C. Nguyen,~\IEEEmembership{Member,~IEEE}

\thanks{Ratun Rahman and Dinh C Nguyen are with the Department of Electrical and Computer Engineering, University of Alabama in Huntsville, Huntsville, AL 35899, USA, emails: rr0110@uah.edu, dinh.nguyen@uah.edu}

}

\markboth{accepted in the IEEE Transactions on Emerging Topics in Computational Intelligence}%
{Shell \MakeLowercase{\textit{et al.}}: Bare Demo of IEEEtran.cls for IEEE Journals}
\maketitle

\begin{abstract}
Conventional federated learning (FL) frameworks often suffer from training degradation due to data uncertainty and heterogeneity across local clients. Probabilistic approaches such as Bayesian neural networks (BNNs) can mitigate this issue by explicitly modeling uncertainty, but they introduce additional runtime, latency, and bandwidth overhead that has rarely been studied in federated settings. To address these challenges, we propose \textbf{Meta-BayFL}, a personalized probabilistic FL method that combines meta-learning with BNNs to improve training under uncertain and heterogeneous data. The framework is characterized by three main features: (1) BNN-based client models incorporate uncertainty across hidden layers to stabilize training on small and noisy datasets, (2) meta-learning with adaptive learning rates enables personalized updates that enhance local training under non-IID conditions, and (3) a unified probabilistic and personalized design improves the robustness of global model aggregation. We provide a theoretical convergence analysis and characterize the upper bound of the global model over communication rounds. In addition, we evaluate computational costs (runtime, latency, and communication) and discuss the feasibility of deployment on resource-constrained devices such as edge nodes and IoT systems. Extensive experiments on CIFAR-10, CIFAR-100, and Tiny-ImageNet show that Meta-BayFL consistently outperforms state-of-the-art methods, including both standard and personalized FL approaches (e.g., pFedMe, Ditto, FedFomo), with up to 7.42\% higher test accuracy. 

\end{abstract}
\IEEEpeerreviewmaketitle
\section{Introduction}
\IEEEPARstart{F}{ederated} \textcolor{black}{learning (FL) has gained considerable momentum due to its ability to facilitate data training at the network edge while ensuring the preservation of privacy \cite{rahman2025electrical2, rahman2024multimodal}.} FL enables local clients to independently train machine learning (ML) models on their local data, subsequently sharing only the model parameters with a centralized server for aggregation without exchanging raw data. This approach significantly improves data privacy and security by keeping local data private within local devices \cite{rahman2024electrical}. Furthermore, FL approaches mitigate communication latency by introducing a decentralized approach and not transmitting only the parameter of local models \cite{nguyen2022latency}. Despite such promising research efforts, \textit{traditional FL frameworks struggle with two practical challenges: uncertain and heterogeneous data}, limiting training performance in federated settings \cite{gawlikowski2023survey}. 

\textbf{1. Uncertain data.} Data in a collaborative environment can be uncertain and the quality of the dataset may vary for each global round and for every client. For example, some client information may be out of date, noisy, or very small in number, which could have a negative impact on the training process when local updates are combined with the global model. This discrepancy increases difficulties in ensuring that the global model remains stable and improves consistently over each training round. Furthermore, uncertainty in data quality can hamper the model's convergence, since the learning algorithm may struggle to establish a generalizable pattern that adequately fits all data quality variations \cite{ye2023adaptive}. This leads to a critical question: a) \textit{How can inherent uncertain data be controlled to stabilize the training of the global model in FL?} 

\textbf{2. Heterogeneous data.} Another significant problem is data heterogeneity where data distributions across clients are naturally non-independent and identically distributed (non-IID). For instance, different clients may have various feature spaces that allow them to be biased towards certain features. These differences create challenges for clients to construct a global model that represents all clients equally, and it can significantly affect the overall effectiveness of FL \cite{t2020personalized,rahman2024improved}. This raises a critical question, b) \textit{How can FL effectively handle heterogeneous data distribution in non-IID settings?}

To address data uncertainty and heterogeneity challenges in FL, several works have been proposed that use advanced learning techniques. Personalized Federated Learning (PFL) has been developed to precisely address heterogeneous data settings. PFL methods ensure that the global model works well across various datasets by tailoring the training process to the features of individual or grouped clients. On the other hand, to address data uncertainty, probabilistic model training methods have been proposed, where probabilistic neural networks use a kernel function to approximate the probability density function of the training data and estimate the probability distribution of the data \cite{specht1990probabilistic}. This allows the model to account for uncertainty in parameters, which is essentially important in training on noisy or unpredictable data. 


\subsection{Related Works} 
We summarize the related works in BNN, probabilistic FL, and personalized FL.

\textbf{1) BNN:} One of the most common probabilistic methods is the bayesian neural network (BNN) which uses the Bayes theorem and observed data to update the model parameters given a prior distribution to estimate the posterior distribution \cite{tran2019bayesian}. BNN utilizes possibilities for all neural network layers, making the prediction more accurate for uncertain data than the basic FL method \cite{doan2025bayesian, jospin2022hands,akram2025uncertainty}.  However, in the BNN, every client is handled equally and as a result, underperforming clients can significantly impact the global model, degrading the training performance. This also leads to poor generalization for individual clients, causing the international model to be biased towards some and irrelevant to others. As a result, the model becomes overly confident and performs poorly, especially on client datasets that differ significantly from the aggregated training data.



\begin{figure}[!t]
\centering
\includegraphics[width=2.5 in]{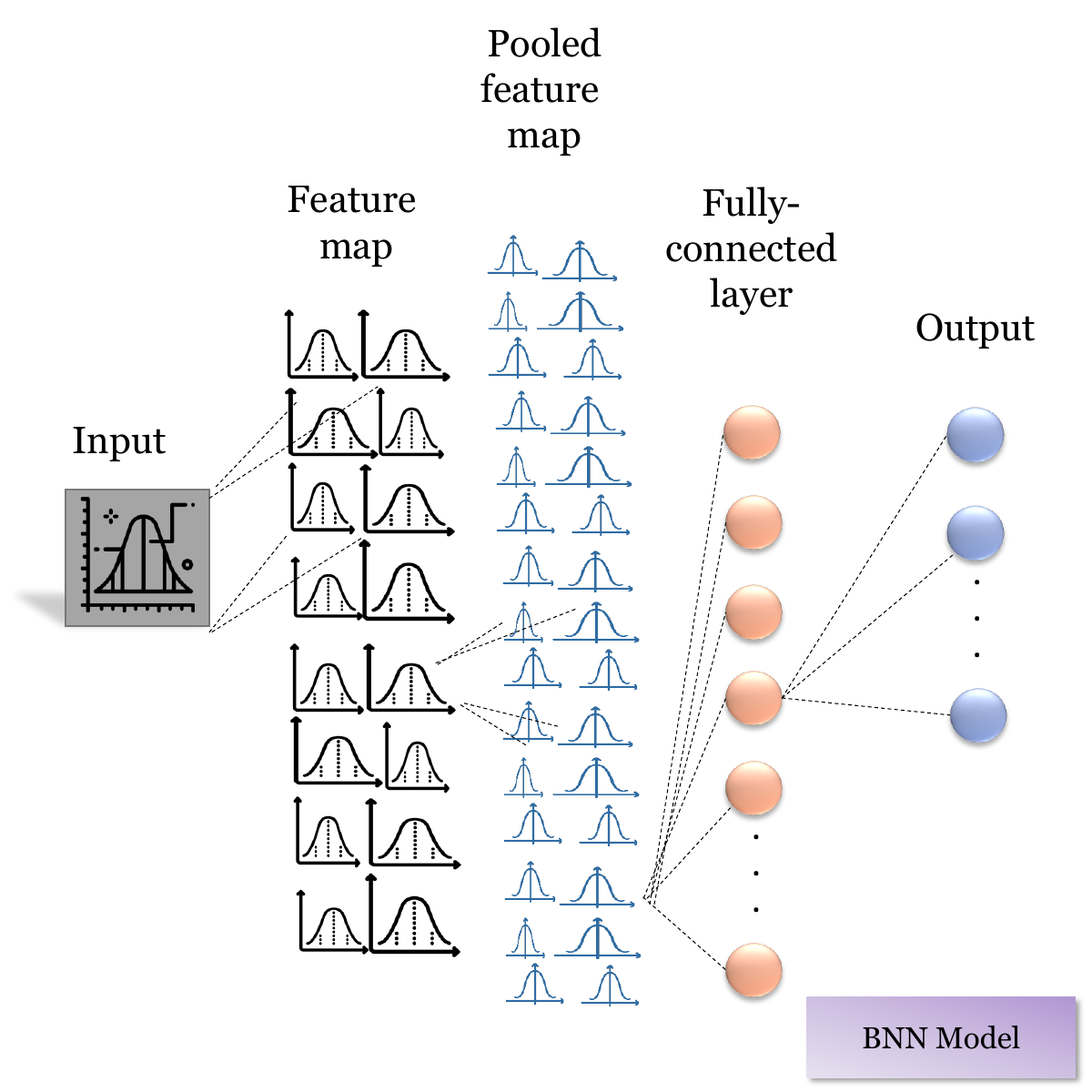}
\caption{\footnotesize \footnotesize \textcolor{black}{A simple probabilistic model structure in which probabilistic properties link every layer to every layer after it. The likelihood graph shows the likelihood of moving from the current layer to the next. It shows that probabilistic (distributional) parameters spread uncertainty through depth.}}
\vspace{-4mm}
\label{Fig: bnn_overview}
\end{figure}

\textbf{2) Personalized FL:} \textcolor{black}{Federated distillation and contrastive learning have also been investigated in recent works to address heterogeneity and communication efficiency. Examples of these include efficient federated distillation for multitask time-series classification \cite{xing2022efficient} and federated contrastive learning with feature-based distillation for human activity recognition \cite{xiao2025federated}. \cite{li2020federated} proposed FedProx that utilizes a proximal term to improve the stability of the FL process. Despite their effectiveness in representation sharing, they do not explicitly describe uncertainty or offer probabilistic personalization under noisy and small client datasets.}

Also methods like pFedMe \cite{t2020personalized} and Ditto \cite{li2021ditto} learned additional personalized models. However, only the desired information that improves the quality of the local model is beneficial for this method, and undesired information leads to poor generalization. Models like Meta-FL \cite{rahman2024improved}, has continuously been used in various fields such as air quality \cite{rahman2025multimodal} and load forecasting \cite{rahman2025electrical}. It uses local personalized aggregation to learn the local model, making the local model more accurate for the global aggregation. However, FedAMP and FedPHP performed their pFL without considering any local objectives. FedMFomo downloaded other client models for aggregation, creating a high communication overhead. 

\textbf{3) Personalized Probabilistic FL: } \textcolor{black}{\cite{li2021fedmask} proposed Fedmask that learns a sparse binary mask for personalized. However, it is mainly used to reduce computational costs and does not enhance the overall performance. pFedBayes \cite{zhang2022personalized} and pFedBL \cite{yu2025pfedbl} also used the pFL method on the Bayesian layer that introduces weight uncertainty for both the clients and server. The computational becomes very expensive in this method for large-scale data. \textit{However, the problem of uncertain and heterogeneous data training in FL settings has been largely under-explored.} It is crucial to develop an efficient model training method to overcome the issue of uncertain and heterogeneous data in real-life applications.}

In our work, we propose \textit{Meta-BayFL}, a personalized probabilistic FL based on meta-learning that focuses on enhancing the training performance by selecting the optimal results. To the best of our knowledge, this is the first work to study meta-learning on probabilistic FL. 


\subsection{Our Key Contributions} Motivated by the above limitations and to further improve the handling of the uncertain data, we propose \textit{Meta-BayFL}, a novel personalized probabilistic FL method with meta-learning-based BNNs on uncertain and heterogeneous data. Our key contributions are summarized as follows:
\begin{itemize}
    \item We introduce a novel notion of model personalization by framing the problem within the probabilistic federated learning (FL) paradigm (Section III). We specifically use meta-learning, a well-established approach for analyzing data in collaborative scenarios with diverse clients. While the structure of FL has been thoroughly researched, the relationship between probabilistic modeling and meta-learning remains largely unexplored. To close this gap, we introduce Meta-BayFL, a personalized probabilistic FL framework that successfully incorporates meta-learning principles. Furthermore, we create an effective computational approach that is designed to meet the problems posed by uncertain and heterogeneous data, improving both adaptability and robustness.

    \item We perform convergence analysis for the proposed \textit{Meta-BayFL} framework (Section IV) where we considered both meta-learning and the probabilistic approach and the derived upper bound reveals the following key properties: 1) using three different learning rates produces three different loss values and using the best learning rate would produce better performance and 2) the loss value decreases when global communication rounds. 

    \item We evaluate our performance with other state-of-the-art methods (Section V). The simulation results show the effectiveness of \textit{Meta-BayFL} for uncertain and heterogeneous data which outperforms other state-of-the-art methods by up to 7.42\% in test accuracy.
\end{itemize}


{\color{black}
\subsection{Paper Structure}
The rest of the paper is organized as follows. Section II discusses the necessary preliminaries, such as Bayesian neural networks, federated learning, and personalized federated learning. The proposed Meta-BayFL framework is described in detail in Section III, along with the training process, the meta-learning-based personalization strategy, and the probabilistic model formulation. A theoretical convergence analysis of Meta-BayFL and upper bounds on the global model performance over communication rounds are presented in Section IV. In-depth experiments on CIFAR-10, CIFAR-100, and Tiny-ImageNet are used in Section V to assess the suggested approach. These experiments include comparisons with cutting-edge federated and personalized learning techniques as well as an examination of runtime, latency, and communication overhead. The paper is finally concluded in Section VI, which also explores possible future research directions.}

{\color{black} \noindent\textbf{Code availability:}
The source code for Meta-BayFL, including all scripts required to reproduce the experimental results, is publicly available at \url{https://github.com/Ratun11/Meta-BayFL}.}

\section{Preliminaries}
\subsection{Bayesian Neural Networks (BNNs)}
Consider the local training iteration index denoted as $t \in \mathcal{T}$,  where $t = \{1, 2, 3, \dots, T\}$ and $\boldsymbol{w}_{n,k}^t$ is the local weight for client $n$ and global round $k$. The local update such as stochastic gradient descent (SGD) at each local iteration $t$ is expressed as: 
\begin{equation} \label{eq: sgd}
\theta^{t+1} = \theta^{t} - \eta \nabla F(\theta,\chi_{n,k}^t),
\end{equation}
where $\eta >0$ is the local learning rate, $\chi$ is the data sample from the local dataset, $F$ is the loss value, and $\theta$ is the local parameter. For BNN, the local model update and aggregation are based on parameter distributions rather than deterministic values in \ref{eq: sgd}. Instead of directly minimizing the loss function $F$, clients aim to maximize their Evidence Lower BOund, $ELBO$ function. Therefore, the local update using BNN is given as
\begin{equation}
    \theta^{t+1} = \theta^t + \eta \nabla ELBO(\theta).
\end{equation}
A higher $ELBO$ indicates a tighter lower bound on the log marginal likelihood, which is defined as:
\begin{equation} \label{eq: elbo}
        ELBO(\theta) = {\mathbb{E}}_{q_\theta} [\log p(D|\theta)] - \text{KL}(q(\theta)||p(\theta)).
\end{equation}
The prior distribution, denoted by $p(\theta)$, represents our assumption and the value of $\theta$ prior to any data being observed and is expressed as
\begin{equation} \label{eq: p}
    p(\theta) = \prod \mathcal{N}(\theta;\mu, \sigma^2),
\end{equation}
where $\mu$ is the mean and $\sigma^2$ is the variance of the distribution. Also, $q(\theta)$ is the approximation to the true posterior distribution $p(\theta|D)$ known as the variational distribution which is defined by
\begin{equation} \label{eq: q}
    q(\theta))= \prod \mathcal{N}(\theta|\mu_v, \sigma_v^2),
\end{equation}
where $\mu_v$ and $\sigma_v^2$ are variational mean and variational variance learned during training. To measure the distance between probability distributions, Kullback–Leibler (KL) divergence is introduced which is given as
\begin{equation} \label{eq: kl}
    KL \left[ q(\theta|D) || p(\theta) \right] = \int q(\theta|D) \log \frac{q(\theta|D)}{p(\theta)} \, d\theta.
\end{equation}
Fig.~\ref{Fig: bnn_overview} describes the probabilistic model architecture where we can see that for each input, we are calculating probabilistic outputs across all the hidden layers. 

\subsection{Federated Learning}
Consider an FL environment where the goal is to minimize the loss value of all the clients, which is usually expressed as
\begin{equation} \label{eq: basicFL}
    \min_{\boldsymbol{w} \in {\mathbb{R}}^d} F(\boldsymbol{w}) := \frac{1}{N} \sum_{n=1}^{N} f_n(\boldsymbol{w}),
\end{equation}
where $\boldsymbol{w}$ represents the model parameter, ${\mathbb{R}}^d$ denotes the $d$-dimensional real space in which the model parameters $\boldsymbol{w}$ reside local loss function, $N$ is the total number of clients participating in federated learning, $f_k(\boldsymbol{w})$ is the local loss function computed by the $i-th$ client using its data. Given the local dataset as $D_n$ the local loss function is denoted by:
\begin{equation}
    f_n(\boldsymbol{w}) = \frac{1}{|D_n|} \sum_{(x_i,y_i)\in D_n} \mathcal{L}(y_{i}, f_{\boldsymbol{w}}(x_i)),
\end{equation}
where $|D_n|$ is the number of data points in $k$ client's dataset and $\mathcal{L}(y_{i}, f_{\boldsymbol{w}}(x_i))$ is the loss for a single data point where $y_i$ is the true label and $f_{\boldsymbol{w}}(x_i)$ is the model's prediction for input $x_i$, parameterized by $\boldsymbol{w}$.

Each client performs SGD locally by iteratively updating its model parameters in the direction that minimizes its own loss function, often represented as:
\begin{equation}
    x_{i,\text{new}} = x_{i,\text{old}} - \eta \nabla f_i(x_{i,\text{old}}),
\end{equation}
where $\eta$ is the learning rate. After performing a certain number of local updates or epochs, each client sends their model updates or gradients to a central server. To improve the global model, the server aggregates these local updates $\boldsymbol{w}_{n,k}^{t}$. The global model at the server is constructed by using the model averaging technique, i.e.,
\begin{equation} \label{eq: fedavg}
    \boldsymbol{w}_{g,k} = \frac{1}{N} \sum_{i=1}^{N} \boldsymbol{w}_{i,k}.
\end{equation}

The central server sends the updated global model parameters back to the clients, and the process repeats for several iterations or until convergence. The server broadcasts the updated global model parameters $\boldsymbol{w}_{\text{global}}$ back to the clients in the next round. 

\subsection{Personalized Federated Learning}
For PFL, every client has their own personalized model where they train their model based on their distinct data distribution. So, the basic FL equation in \ref{eq: basicFL} becomes
\begin{equation} \label{eq: PFL}
    \min_{x \in {\mathbb{R}}^d} F(x) := \frac{1}{|\mathcal{N}|} \sum_{n=1}^{|\mathcal{N}|} f_i(x) + x_n, 
\end{equation}
where $x_n$ is the personalized term for client $n\in \mathcal{N}$. The local SGD update can be expressed as:
\begin{equation}
\boldsymbol{w}_{n,k}^{t+1} = \boldsymbol{w}_{n,k}^{t} - \eta_k \nabla F(\boldsymbol{w}_{n,k}^t,\chi_{n,k}^t) + \textbf{v}_{n,k},
\end{equation}
where $\chi$ is the local dataset non-IID sample, and $\textbf{v}_{n,k}$ is the $n$ client's personalized value for global round $k$. Each client updates its personalized model as follows:
\begin{equation}
\boldsymbol{x}_{n,k}^{t+1} = \boldsymbol{x}_{n,k}^{t} - \eta_k\left[\nabla F(\boldsymbol{x}_{n,k}^t,\zeta_{n,k}^t) + \lambda(\boldsymbol{x}_{n,k}^{t} -\boldsymbol{w}_{n,k}^{t}) \right],
\end{equation}
where $\eta$ is the personalized learning rate, $\zeta$ is a local data sample, and $\lambda >0$ is the parameter that controls the interpolation of global and personalized models. Federated averaging is carried out at the server like \ref{eq: fedavg}.  

\section{Proposed Probabilistic FL Framework with Model Personalization}

\begin{figure}[!t]
\centering
\includegraphics[width=3.5 in]{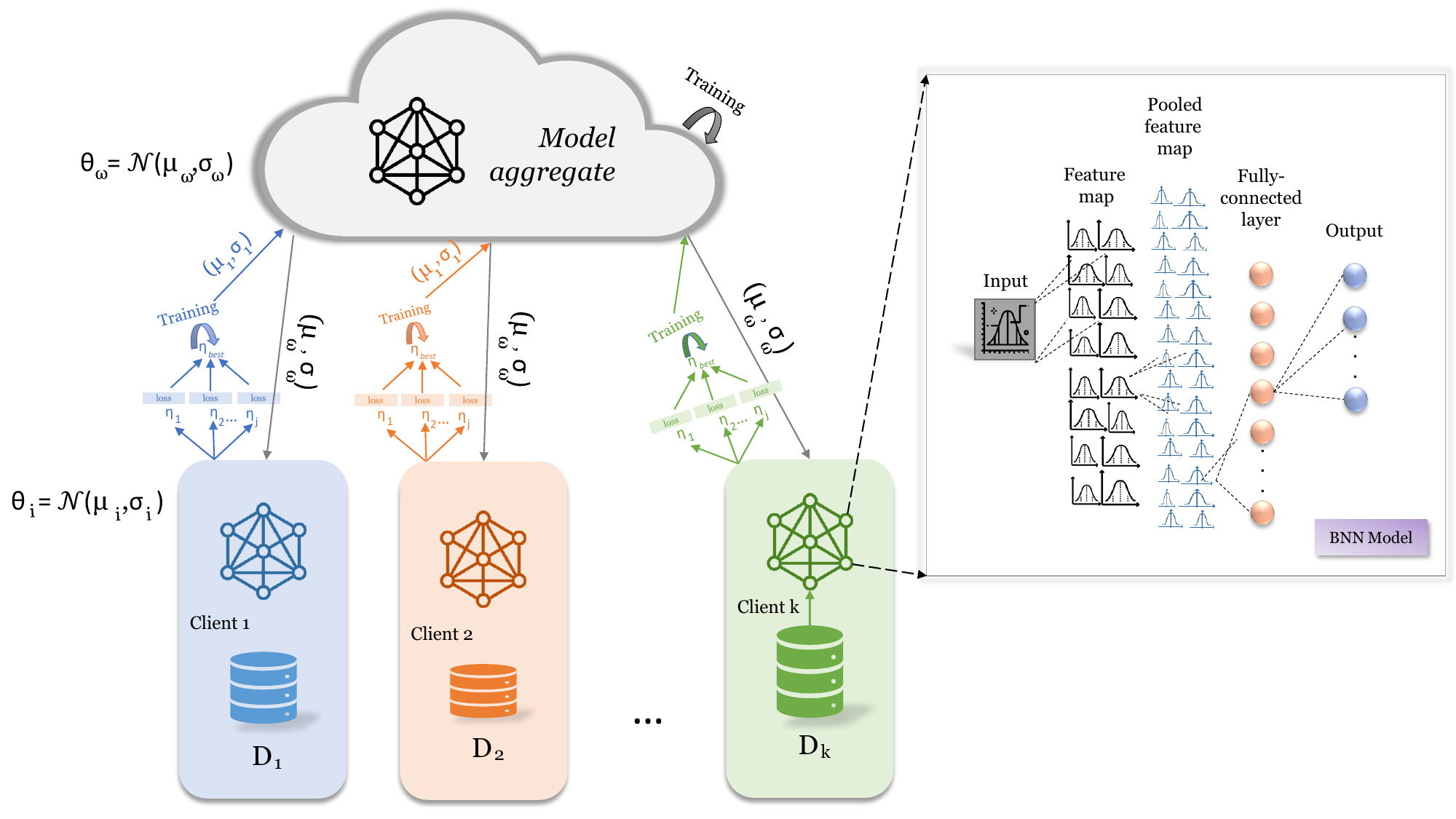}
\caption{\footnotesize \textcolor{black}{Our proposed \textit{Meta-BayFL} FL algorithm where there are $K$ clients connected to the global server with adaptive model aggregation using a personalized learning approach. Each client creates local models using BNN and their local data and different temporary learning rates, and chooses the best rates for local model training. The server aggregates the local models and distributes the updated global model to all the clients.}}
\vspace{-1mm}
\label{Fig: bnn_overview2}
\end{figure}

We consider an FL system where a set of clients denoted by as $\mathcal{N}$ collaborate to train a shared ML model with a single server, as shown in Fig. \ref{Fig: bnn_overview2}. For each global round $k \in \mathcal{K}$, each client $n \in \mathcal{N}$ trains a BBN model represented by $\boldsymbol{w}_{n,k}^t$ where local BNN iteration is denoted as $t \in \mathcal{T}$. 
\subsection{Model Training Concept}
Assuming the number of available learning rates as $j$, we can separate our system model into multiple steps as follows.

\textbf{Step 1:}
In every $k$, the global model's weight $(\boldsymbol{w}_{g,k})$ is distributed across all $n$. All $n$ receives $\boldsymbol{w}_{g,k}$ as well as their local dataset denoted by $D_n^k$ for local model raining. We denote local model's weight as $\boldsymbol{w}_{n,k}$, and so initially, $\boldsymbol{w}_{n,k} = \boldsymbol{w}_{g,k}$.

\textbf{Step 2:}
Every $n$ creates a small testing dataset from $D_n^k$ denoted as $d_n^k \subset D_n^k$ and is used to build the optimal local model using meta-learning. 

\textbf{Step 3:}
Under the federated setting,  the prior $p(\boldsymbol{w}_{n,k}^t)$, posterior $q(\boldsymbol{w}_{n,k}^t)$ distribution, and KL divergence can be expressed as

\begin{equation} \label{eq: updated p}
    p(\boldsymbol{w}_{n,k}^t) = \prod \mathcal{N}(\boldsymbol{w}_{n,k}^t;\mu, \sigma^2),
\end{equation}
\begin{equation} \label{eq: updated q}
    q(\boldsymbol{w}_{n,k}^t)= \prod \mathcal{N}(\boldsymbol{w}_{n,k}^t|\mu_v, \sigma_v^2),
\end{equation}
and
\begin{equation} \label{eq: updated kl}
\begin{split}
    KL \left[ q(\boldsymbol{w}_{n,k}^t|d_n^k) || p(\boldsymbol{w}_{n,k}^t) \right] \\ 
    = \int q(\boldsymbol{w}_{n,k}^t|d_n^k) \log \frac{q(\boldsymbol{w}_{n,k}^t|d_n^k)}{p(\theta)} \, d\boldsymbol{w}_{n,k}^t.
\end{split}
\end{equation}

\textbf{Step 4:}
We calculate $ELBO$ using equation \ref{eq: elbo} as follows.
\begin{equation} \label{eq: updatedelbo}
\begin{split}
        ELBO(\boldsymbol{w}_{n,k}^t) = {\mathbb{E}}_{q(\boldsymbol{w}_{n,k}^t)} [\log p(d_n^k|\boldsymbol{w}_{n,k}^t)] \\ 
        - \text{KL}[q(\boldsymbol{w}_{n,k}^t)||p(\boldsymbol{w}_{n,k}^t)].
\end{split}
\end{equation}

\textbf{Step 5:}
The local model is updated for every temporary local round $t_{\text{temp}}$ using different learning rates $\eta_i$ where $i \in j$ for all $j$ as follows.
\begin{equation}
    \boldsymbol{w}_{n,k}^{t_{\text{temp}}+1} = \boldsymbol{w}_{n,k}^{t_{\text{temp}}} + \eta_i \nabla ELBO(\boldsymbol{w}_{n,k}^{t_{\text{temp}}}).
\end{equation}
After $t_{\text{temp}}$ temporary local rounds, we get a temporary local model for learning rate $\eta_i$. Let us denote the temporary local model as $\boldsymbol{w}_{n,k}^i$. For all learning rates $j$, we will get the total $j$ number of temporary local models in every $n$ and every $k$. 

\textbf{Step 6:}
After calculating all the temporary local models, we calculate the temporary loss value based on those models on dataset $d_n^k$ as follows
\begin{equation} \label{eq: temploss}
    f_{n,k}(\boldsymbol{w}_{n,k}^i) = \frac{1}{|d_n^k|} \sum_{(x_i,y_i)\in d_n^k} \mathcal{L}(y_{i}, f_{\boldsymbol{w}_{n,k}^i}(x_i)).
\end{equation}
We calculate the loss value for every learning rate $\eta_i$ where $i \in j$. The learning rate that provides the lowest $f_n$ value is considered the optimal learning rate denoted by $\eta^{best}_{n,k}$ for the local model training, i.e., 
\begin{equation} \label{eq: beta}
    \eta^{best}_{n,k} := \argmin_{i \in j} f_{n,k} (\boldsymbol{w}_{n,k}^i). 
\end{equation}
\textbf{Step 7:}
We use $\eta^{best}_{n,k}$ for the local model training on client $n$ and global round $k$. So for local round $t$,
\begin{equation}
    \boldsymbol{w}_{n,k}^{t+1} = \boldsymbol{w}_{n,k}^t + \eta^{best}_{n,k} \nabla ELBO(\boldsymbol{w}_{n,k}^t).
\end{equation}
After $T$ local epochs, we finally get the updated local model $\boldsymbol{w}_{n,k} = \boldsymbol{w}_{n,k}^T$. The local model is sent to the server.

\textbf{Step 8:}
The server collects all the local models' weights for the global model aggregation as follows:
\begin{equation} \label{eq: updatedfedavg}
    \boldsymbol{w}_{g,k+1} = \frac{1}{N} \sum_{i=1}^{N} \boldsymbol{w}_{i,k}.
\end{equation}
Finally, the updated global model $\boldsymbol{w}_{g,k+1}$ is distributed across all clients $k$ for the $K+1$ global round. After $K$ global rounds, we finally get the optimal global model $\boldsymbol{w}^*$.

In our approach, we used 80\% of our data for local model training and expressed them as Trainloader $(D_n^k)$ and the rest 20\% for testing as Testloader $(D_n^{'k})$. To evaluate the performance, we used 20\% of the training data expressed as Dataloader, $d_n^k$ where $d_n^k \subset D_n^k$ and used to find the optimal learning rate $\eta^{best}_{n,k}$.
\begin{algorithm}
\footnotesize
	\caption{\footnotesize Proposed  FL algorithm}
	\begin{algorithmic}[1]
		\label{algo: metaSGD}
		\STATE \textbf{Input:}  The set of global communication rounds $\mathcal{K}$, local training round $\mathcal{T}$, number of clients $\mathcal{N}$, set of available learning rates $\eta_{1,2,...,j}$, the temporary round for meta-learning as $T_{\text{temp}}$
		\STATE \textbf{Initialization:} Initialize global model $\boldsymbol{w}_0$
		\FOR{each global communication round $k \in \mathcal{K}$}
		\STATE Send $\boldsymbol{w}_{g,k}$ to $\mathcal{N}$ sampled clients
		\FOR{each sampled client $n \in \mathcal{N}$}
        \STATE Receive the global weight as local weight $\boldsymbol{w}_{n,k} = \boldsymbol{w}_{g,k}$
		\FOR{each local training epoch $t \in \mathcal{T}$}
        \STATE Compute the prior $p(\boldsymbol{w}_{n,k})$ distribution, posterior $q(\boldsymbol{w}_{n,k})$ distribution, and KL divergence using equation \ref{eq: updated p}, \ref{eq: updated q}, and \ref{eq: updated kl}
        \STATE Calculate $ELBO$ using the variables above and equation \ref{eq: updatedelbo}
        
        \FOR{each learning rates $\eta_j$ where $i \in j$}
        \STATE Update the temporary local model weight using $\eta_i$ for all local temp rounds $t_{\text{temp}}$ as $\boldsymbol{w}_{n,k}^i$
        \STATE Perform test on $\boldsymbol{w}_{n,k}^i$ using equation \ref{eq: temploss} on data $d_n^k$
        \STATE Calculate $loss(\eta_j) = f_i(w, D_n^{'k})$ on $D_n^{'k}$
        \STATE Save the $\eta_i$ that has the lowest $loss$ value as $\eta^{best}_{n,k}$ using equation \ref{eq: beta}
        \STATE Return $\eta^{best}_{n,k}$
        \ENDFOR
        
        \STATE Aggregate the local model on $D_n^k$ using $\eta^{best}_{n,k}$, $\boldsymbol{w}_{n,k}^{t+1} \longleftarrow \boldsymbol{w}_{n,k}^t + \eta^{best}_{n,k} \nabla ELBO(\boldsymbol{w}_{n,k}^t)$
		\ENDFOR
        \STATE Send $\boldsymbol{w}_{n,K} = \boldsymbol{w}_{n,K}^{T}$ to server 
		\ENDFOR
		\STATE The server updates the global parameter by averaging: $\boldsymbol{w}_{g,k+1} = \frac{1}{N} \sum_{n\in\mathcal{N}}\boldsymbol{w}_{n,k}$ 
		\STATE The server broadcasts it to all clients for the next round of training
		\ENDFOR
		\STATE \textbf{Output:} Optimal global model $\boldsymbol{w}^*$
	\end{algorithmic}
\end{algorithm} 

The proposed PFL approach is summarized in Algorithm~\ref{algo: metaSGD}. Here, for every global round $k$, we send global weight to all clients in line 4. Then we calculate prior distribution, posterior distribution, KL divergence, and ELBO in every local round $t$ explained in lines 8-9. Lines 10-16 focus on the meta-learning functionalities where total $j$ learning rates ($\eta_i$ where $i \in j$) are used on $d_n^k$ before local model aggregation. The learning rate that produces the lowest loss value is used to aggregate the local model in line 17. We send the optimal local weight to the server in line 19. The server collects all the local weights from every client $n$ calculates federated averaging in line 21 and saves the updated global weight for the next global rounds. \textcolor{black}{Among the three learning rate schedules analyzed in Theorem~1, we adopt the meta-learned adaptive learning rate in all experiments, as it empirically yields faster convergence and improved stability under non-IID and noisy settings. The remaining schedules are included in the analysis to establish general convergence guarantees and to serve as theoretical baselines.}

\subsection{Convergence Analysis}


To support our convergence analysis, we introduce two virtual sequences:
\begin{equation}
\bar{\boldsymbol{w}}_k^t = \frac{1}{N} \sum_{n\in\mathcal{N}} \boldsymbol{w}_{k}^t,~~~~~~~~~~~~~~~~~~~~\bar{\boldsymbol{x}}_k^t = \frac{1}{N} \sum_{n\in\mathcal{N}} \boldsymbol{x}_{k}^t.
\end{equation}

Subsequently, each client  updates its personalized model as
\begin{equation}
\boldsymbol{x}_{k}^{t+1} = \boldsymbol{x}_{k}^{t} - \eta_t g_t,~~~~~~~~~~~~~~~~~~~~g_t=\nabla f(\boldsymbol{x}_{k}^{t}) + b_t + n_t
\end{equation}
where for zero-mean noise $\mathbb{E} n_t=0$ and bias $b_k$, $g_t$ is a  gradient oracle and $\eta_t$ is the sequence of step sizes. If there is no bias, $b_t=0$, it becomes the SGD setting and for no noise, $n_t=0$, it becomes the classic gradient descent algorithm. 

It is easy to observe that,
\begin{equation}
\label{eq:fl_short}
    \bar{\boldsymbol{w}}_{k}^{t+1} = \boldsymbol{w}_{k}^{t} - \eta_k \nabla F(\boldsymbol{w}_{k}^t,\chi_{k}^t) + \eta_k B_k - \eta_k N_k,
\end{equation}

 To facilitate the analysis, we use the following common assumptions:
 \begin{assumption} \label{assump:1} ($L-$smoothness).
 Each local loss function $F_n$ ($n\in \mathcal{N}$) is $L$-smooth ($L>0$), i.e.
 \begin{equation}
F_n(\boldsymbol{w}') - F_n(\boldsymbol{w}) \leq \langle \boldsymbol{w}'- \boldsymbol{w}, \nabla F(\boldsymbol{w} \rangle + \frac{L}{2} ||\boldsymbol{w}'- \boldsymbol{w}||, \forall \boldsymbol{w}', \boldsymbol{w}
 \end{equation}
 \end{assumption}

 \begin{assumption} \label{assump:2} (($M, \sigma^2$)-bounded noise).
     There exists constant $M$, $\sigma^2 >= 0$ such that
     \begin{equation}
         \mathbb{E} || n(\boldsymbol{w},\xi)||^2 \leq M ||\nabla F_n(\boldsymbol{w}) + b(\boldsymbol{w})||^2 + \sigma^2, \forall \boldsymbol{w} \in \mathbb{R}^d.
     \end{equation}
 \end{assumption}

\begin{assumption} \label{assump:3} (($m, \zeta$)-bounded bias).
    There exists constants $0 \leq m < 1$ and $\zeta^2 \geq 0$ such that
    \begin{equation}
        ||b(\boldsymbol{w}||^2 \leq m||\nabla F_n(\boldsymbol{w})||^2 + \zeta^2, \forall\boldsymbol{w} \in \mathbb{R}^d.
    \end{equation}
\end{assumption}

\begin{assumption} \label{assump:4} Finite parameter space: The parameter space $\Theta$ is finite, i.e, $\Theta = {\chi_{k}^1,\chi_{k}^2,\chi_{k}^3,\dots,\chi_{k}^t}$. Also, $\chi_{k}^{'t}\in\Theta$

There exist $0<L_H<\infty$ such that 
\begin{equation}
\begin{aligned}
    ||\nabla F(\boldsymbol{w},\chi_{k}^1)-\nabla F(\boldsymbol{w},\chi_{k}^2)||_2 &\leq L_H||\chi_{k}^1-\chi_{k}^2||_2 \forall \chi_{k}^1,\chi_{k}^2 \in \Theta, \\&\forall x\in \mathscr{X}
\end{aligned}
\end{equation}

Sampling variance is bounded by $\sigma^2$, such that
\begin{equation}
    \mathbb{E}[||\nabla f(\boldsymbol{w}*,\xi)-\nabla F(\boldsymbol{w}_{k}^t,\chi_{k}^t)||_2^2|\chi_n] \leq \sigma^2, \forall \chi_{k}\in \Theta
\end{equation}
\end{assumption}
\begin{assumption} \label{Assump:Variance-gradient}
 The variance of stochastic gradients on local model training at each client is bounded: $\mathbb{E}||\nabla F(\boldsymbol{w}_{n,k}^t,\chi_{n,k}^t) - \nabla F(\boldsymbol{w}_{n,k}^t)||^2 \leq \sigma_g^2$
 \end{assumption}
\begin{lemma} \label{lemma1}
    Under Assumption \ref{assump:4}, there exist a constant $C_1 > 0$ such that for any $\delta > 0$, with probability $1-\delta$ we have
    \begin{equation}
        \begin{aligned}
            ||\mathbb{E}_{\pi_t} \nabla F(\boldsymbol{w}_{k},\chi_{k}^t) -\mathbb{E}_{\pi_1} \nabla F(\boldsymbol{w}_{k},\chi_{k}^{'t})||_2^2 \leq C_1 \frac{log Dt + log\frac{1}{\delta}}{Dt},
        \end{aligned}
    \end{equation}
\end{lemma}
$\forall x\in \mathscr{X}, \forall t>0$

\begin{lemma} \label{lemma2}
Let F be L-smooth, $\boldsymbol{x}_{k}^{t+1}$ and $\boldsymbol{x}_{k}^{t}$ as in \ref{eq:fl_short} with Assumption \ref{assump:2} and \ref{assump:3}. For any stepsize $\eta \leq \frac{1}{(M+1)L}$, it holds
\begin{equation} 
\begin{aligned}
\mathbb{E}_\xi [F(\boldsymbol{w_{k}^{t+1}}) - F(\boldsymbol{w_{k}^t}) | \boldsymbol{w_{k}^t}] \leq &\frac{\eta(m-1)}{2} ||\nabla F(\boldsymbol{w_{k}^t})||^2 + \\&\frac{\eta}{2} \zeta^2+ \frac{\eta^2L}{2} \sigma^2
\end{aligned}
\end{equation}
when $M=m=\zeta^2=c$ for any constant $c$, we recover the standard descent lemma. 
\end{lemma}

\begin{lemma} \label{lemma3}
    Under Assumption \ref{assump:4}, there exists a constant $C_1>0$ such that for any $\delta>0$, with probability at least $1-\delta$ we have

    \begin{equation}
    \begin{aligned}
        ||\mathbb{E}\nabla F(\boldsymbol{w},\chi^t) - &\mathbb{E}_{\pi_1} \nabla F(\boldsymbol{w},\chi^{'t})||_2^2 \leq C_1\frac{log Dt+log \frac{1}{\delta}}{Dt}, \\&\forall x\in\mathscr{X}, \forall t>0.
    \end{aligned}
    \end{equation}
\end{lemma}

\begin{lemma} \label{lemma4}
    Under the assumption $p = P_s$ and series $\sum_{l\geq1}N_l e^{-l^2}$ converges. 

    \begin{equation}
        \bar v[B(s,\frac{k}{\sqrt{n}})|X] \geq 1 - \delta, \forall s \in S,  \epsilon,\delta \in (0,1)
    \end{equation}
    with probability at least $1-\epsilon$ with respect to $P_s^n$, and where $k=k(\epsilon,\delta,v(s)) = inf \left[j \geq 1|\sum_{l\geq j}N_l e^{-l^2} \leq \epsilon \sqrt{\delta v(s)}\right]$ is independent of $n$ and nonincreasing with the positive parameters $\epsilon,\delta, and v(s)$.
\end{lemma}
\begin{lemma} \label{lemma5}
Let Assumption \ref{Assump:Variance-gradient} hold, the expected upper bound of the variance of the stochastic gradient on local model training is given as 
\begin{equation}
\mathbb{E} ||g_k^t- \bar{g}_k^t||^2 \leq \frac{\sigma_g^2}{N^2}.
\end{equation}
\end{lemma}

\begin{lemma} \label{lemma6}
The expected upper bound of the divergence of $\boldsymbol{w}_{n,k}^t$ is given as 
\begin{equation} 
\begin{aligned}
& \left[ \frac{1}{N}\sum_{n\in\mathcal{N}}\mathbb{E} \Big\Vert\bar{\boldsymbol{w}}_k^t-\boldsymbol{w}_{n,k}^t\Big\Vert^2  \right] \leq  4\eta_kT B^2,
\end{aligned}
\end{equation}
for some positive $B$.
\end{lemma}

\begin{lemma} \label{lemma7}
    The expected upper bound of $\mathbb{E}[||\bar{\boldsymbol{w}_k^{t+1}}-\boldsymbol{w}^*||^2]$ is given as
    \begin{equation} 
\begin{aligned}
\mathbb{E}||\bar{\boldsymbol{w}}_k^{t+1} - \boldsymbol{w}^*||^2  
&\leq ||(1-\mu\eta_k)||\bar{\boldsymbol{w}}_k^t - \boldsymbol{w}^*||^2  + \frac{1}{N}\sum_{n\in\mathcal{N}}||\bar{\boldsymbol{w}}_k^t-\\&\boldsymbol{w}_{n,k}^t||^2 + 
\frac{1}{4\eta_k}\frac{1}{N} \sum_{n\in\mathcal{N}}||\boldsymbol{w}_{n,k}^t -  \bar{\boldsymbol{w}}_k^t||^2 + \\&\frac{1}{2} \min (\Theta_1,\Theta_2,\Theta_3)  + \eta_k^2||g_k^t - \bar{g}_k^t||^2
\end{aligned}
\end{equation}
\end{lemma}

\begin{theorem} \label{theorem2}
    Under some Assumptions, for any $\delta > 0$, we have the probability at least $1-\delta$, for any $T > 0$, the following bound on the expected gradient of the final output under the true parameter $\chi_{k}^{'t}$
    
    \begin{enumerate}
        \item[(i)] If the step size ($\eta$) satisfies $\eta_k = \frac{a}{\sqrt{K}}, \forall k \leq \mathcal{K}$, for some constant $a < \frac{\sqrt{\mathcal{K}}}{L_h}$, then 
        
        $\mathbb{E}[||\nabla F(z_{\mathcal{K}},\chi_{k}^{'t})||_2^2] \leq \bigg[\frac{2(F(\boldsymbol{w}_1,\chi_{k}^{'t})- \min_{x\in \mathscr{X}}F(\boldsymbol{w},\chi_{k}^{'t}))}{a\sqrt{\mathcal{K}}}\bigg]+\bigg[\frac{A_1}{\mathcal{K}}+\frac{A_2 log \mathcal{K}}{\mathcal{K}}+\frac{A_3 log^2\mathcal{K}}{\mathcal{K}}\bigg]+\frac{L_ha\sigma^2}{\sqrt{\mathcal{K}}}$,
        
        where $A_1 = \frac{C_1(log D-log \delta)}{L_h D}, A_2=\frac{C_1(log D-log \delta)}{L_h D}+\frac{C_1}{L_hD}, A_3=\frac{C_1}{L_h D}$. \\\\

        \item[(ii)] If the step size ($\eta$) satisfies $\eta_k = \frac{a}{k}, \forall k \leq \mathcal{K}$, for some constant $a < \frac{1}{L_h}$, then 
        
        $\mathbb{E}[||\nabla F(z_{\mathcal{K}},\chi_{k}^{'t})||_2^2] \leq \bigg[\frac{2(F(\boldsymbol{w}_1,\chi_{k}^{'t})- \min_{x\in \mathscr{X}}F(\boldsymbol{w},\chi_{k}^{'t}))}{a} +\frac{6C_1+\pi^2C_1(log D - log \delta)}{6D} +\frac{\pi^2L_h a\sigma^2}{6}\bigg] \frac{1}{log \mathcal{K}}$. \\\\

        \item[(iii)] if the step size ($\eta$) satisfies $\eta_k = \frac{a}{\sqrt{k}}, \forall k \leq \mathcal{K}$, for some constant $a < \frac{1}{L_h}$, then 
        
        $\mathbb{E}[||\nabla F(z_{\mathcal{K}},\chi_{k}^t)||_2^2] \leq \bigg[\frac{2(F(\boldsymbol{w}_1,\chi_{k}^{'t})- \min_{x\in \mathscr{X}}F(\boldsymbol{w},\chi_{k}^{'t}))}{a\sqrt{\mathcal{K}}} + \frac{3C_1(log D - log \delta) +4C_1}{D\sqrt{\mathcal{K}}}+\frac{L_ha\sigma^2}{\sqrt{\mathcal{K}}}\bigg]+\frac{L_ha\sigma^2log \mathcal{K}}{\sqrt{\mathcal{K}}}.$
    
    \end{enumerate}
\end{theorem}

\begin{remark}
    Different $\eta$ values produced different loss values as shown in Theorem \ref{theorem2}. The theorem shows that in (i) the $\eta$ value $\frac{a}{\sqrt{K}}$ for constant $a < \frac{\sqrt{\mathcal{K}}}{L_h}$, (ii) the $\eta$ value $\eta_k = \frac{a}{k}$ for constant $a < \frac{1}{L_h}$, and (iii) the $\eta_k = \frac{a}{\sqrt{k}}$ for constant $a < \frac{1}{L_h}$ has produces three different $\mathbb{E}[||\nabla F(z_{\mathcal{K}},\chi_{k}^{'t})||_2^2]$ values. In basic FL, there are only one $\eta$ value and that produce one $\mathbb{E}[||\nabla F(z_{\mathcal{K}},\chi_{k}^{'t})||_2^2]$ value. 
\end{remark}
\begin{theorem} \label{theorem1}
Based on the above Lemmas and Theorem \ref{theorem2}, the convergence bound of our approach after $K$ global communication rounds is given as 
\begin{equation} \label{equa:final_convergenceFunction}
\begin{aligned}
    &\mathbb{E}\left[F(\boldsymbol{w}_K)\right] -F^* \\&\leq \frac{L}{2(K+L/\mu)}\left[\frac{16\Phi_K}{15\mu^2} + \left( \frac{L}{\mu}+1\right) \mathbb{E}||\boldsymbol{w}_0 - \boldsymbol{w}^*||^2 \right].
\end{aligned}
\end{equation}
\label{theorem_final}
\end{theorem}

\begin{remark}
    The upper bound of convergence achieved in Theorem \ref{theorem1} shows that the convergence of the \textit{Meta-BayFL} algorithm is strongly influenced by the number of total communication rounds. 
\end{remark}

\section{Detailed Proofs of Convergence}

\subsection{Proof of Lemma \ref{lemma2}}
By the quadratic upper bound in Assumption \ref{assump:1} and Assumption \ref{assump:2}:
\begin{equation*}
    \begin{aligned}
        &\mathbb{E} F(\boldsymbol{w}_{k}^{t+1}) \\&\leq F(\boldsymbol{w}_{k}^t) -\eta_k(\nabla F(\boldsymbol{w}_{k}^t), \mathbb{E} g_t) + \frac{\eta^2 L}{2} (\mathbb{E} || g_t - \mathbb{E} g_t||^2 + \mathbb{E} \\&~~~~||\mathbb{E} g_t||^2)
        \\&= F(\boldsymbol{w}_{k}^t) -\eta_k(\nabla F(\boldsymbol{w}_{k}^t), \nabla f(\boldsymbol{x}_{k}^{t}) + b_t + n_t) +\\&~~~~ \frac{\eta^2 L}{2} (\mathbb{E} || n_t||^2 + \mathbb{E} ||\nabla f(\boldsymbol{x}_{k}^{t}) + b_t + n_t||^2)
        \\&\leq F(\boldsymbol{w}_{k}^t) -\eta_k(\nabla F(\boldsymbol{w}_{k}^t), \nabla f(\boldsymbol{x}_{k}^{t}) + b_t + n_t) + \frac{\eta^2 L}{2} ((M+1) \\&~~~~\mathbb{E} ||\nabla f(\boldsymbol{x}_{k}^{t}) + b_t + n_t||^2 + \sigma^2)
    \end{aligned}
\end{equation*}

By the choice of stepsize, $\eta\leq \frac{1}{(M+1)L}$, and Assumption \ref{assump:3}:
\begin{equation}
    \begin{aligned}
        &\mathbb{E} F(\boldsymbol{w}_{k}^{t+1}) \\&\leq F(\boldsymbol{w}_{k}^t) +\frac{\eta_k}{2}(-2(\nabla F(\boldsymbol{w}_{k}^t), \nabla f(\boldsymbol{x}_{k}^{t}) + b_t + n_t) +  \\&~~~~||\nabla f(\boldsymbol{x}_{k}^{t}) + b_t + n_t||^2) + \frac{\eta^2 L}{2} \sigma^2
        \\&= F(\boldsymbol{w}_{k}^t) +\frac{\eta_k}{2}(-||\nabla F(\boldsymbol{w}_{k}^t||^2 + ||b_t+n_t||^2) + \frac{\eta^2 L}{2} \sigma^2
        \\&= F(\boldsymbol{w}_{k}^t) +\frac{\eta_k}{2}(m-1)||\nabla F(\boldsymbol{w}_{k}^t)||^2 + \frac{\eta}{2}\zeta^2 +\frac{\eta^2 L}{2} \sigma^2
    \end{aligned}
\end{equation}
This concludes the proof. 

\subsection{Proof of Lemma \ref{lemma3}} The Hellignger distance between $\theta_1$ and $\theta_2$ 
\begin{equation}
    d(\theta_1,\theta_2) = \sqrt{\frac{1}{2}\int_{\mathscr{Y}}(\sqrt{f(y;\theta_1}-\sqrt{f(y;\theta_2)})^2}
\end{equation}
There exists a constant A such that $||\theta_1=\theta_2|| \leq A d(\theta_1,\theta_2)$, where $||.||$ is the Euclidean norm. Let $B_k^t = B(\theta^c, \frac{k}{\sqrt{Dt}})$ be a ball centered at $\theta^c$ with radius $\frac{k}{\sqrt{Dt}}$ under distance $d$. Since $\Theta$ is finite, we can directly apply Lemma \ref{lemma4}. So for $t\leq T, \epsilon, \delta \in (0,1)$ with probability at least $1-\frac{6\delta}{\pi^2 t^2}$ with respect to $\mathbb{P}_{\theta^c}^t$, we have

\begin{equation}
    \pi _t(B_{k(t)}^t) \geq 1 - \epsilon,
\end{equation}
where $k(t) = inf \left[j \geq 1 | \sum_{i\geq j}|\Theta|e^{-i^2} \leq \frac{6\delta}{\pi^2 t^2} \sqrt{\epsilon \pi_0 (\theta^c)}\right]$.

Note that $\sum_{i\geq j} e^{-i^2} \leq \frac{e}{e-1} e^{-j^2}$, we can set k(t) to be the solution of next equation. 
\begin{equation}
    \frac{e}{e-1} |\Theta|e^{-k(t)^2} = \frac{6\delta}{\pi^2 t^2} \sqrt{\epsilon \pi_0 (\theta^c)}
\end{equation}

we get $k(t) = \sqrt{log \frac{e|\Theta|\pi^2 t^2}{6\delta (e-1) \sqrt{\epsilon, \pi_0 (\theta^c)}}}$. Now the bias in the gradient estimator can be bonded as follows.

\begin{equation}
\footnotesize
    \begin{aligned}
        &||\mathbb{E}_{\pi_t} \nabla_x F(\boldsymbol{w},\chi) - \mathbb{E}_{\pi_t} \nabla_x F(\boldsymbol{w},\chi^')||_2^2 
        \\& = ||\int(\nabla_x F(\boldsymbol{w},\chi) - \nabla_xF(\boldsymbol{w},\chi^'))\pi_t (\theta) d\theta||_2^2
        \\& \leq \int||(\nabla_x F(\boldsymbol{w},\chi))-(\nabla_x F(\boldsymbol{w},\chi^'))_2^2||\pi_t (\theta) d\theta
        \\&\leq L_H^2||\chi-\chi^'||_2^2 \pi_t (\theta) d\theta
        \\&= F(\boldsymbol{w}_{k}^t) -\eta_k(\nabla F(\boldsymbol{w}_{k}^t), \nabla f(\boldsymbol{x}_{k}^{t}) + b_t + n_t) + \frac{\eta^2 L}{2} (\mathbb{E} || n_t||^2 + \\&~~~~\mathbb{E} ||\nabla f(\boldsymbol{x}_{k}^{t}) + b_t + n_t||^2)
        \\& = \int_{B_{k(t)}^t} L_H^2||\chi-\chi^'||_2^2 \pi_t (\theta) d\theta + \int_{(B_{k(t)}^t)^'} L_H^2||\chi-\chi^'||_2^2 \pi_t (\theta) d\theta
        \\&\leq A^2 L_H^2 \frac{k(t)^2}{Dt} \int_{B_{k(t)}^t}\pi_t(\theta) d\theta + L_H^2 \max_{\chi \in \Theta}||\chi-\chi^'||_2^2\int_{(B_{k(t)}^t)^'} \pi_t (\theta) d\theta
        \\& \leq A^2 L_H^2 \frac{k(t)^2}{Dt}+ L_H^2 \max_{\chi \in \Theta}||\chi-\chi^'||_2^2 \epsilon
    \end{aligned}
\end{equation}

Here, D is the data batch size. Note that $\epsilon = \frac{1}{Dt}$ and $k(t) = \sqrt{log\frac{e|\Theta|\pi^2t^2\sqrt{Dt}}{6\delta(e-1)\sqrt{\pi_0(\chi')}}}$, we have
\begin{equation}
    \begin{aligned}
        &||\mathbb{E}_{\pi_t} \nabla_x F(\boldsymbol{w},\chi) - \mathbb{E}_{\pi_t} \nabla_x F(\boldsymbol{w},\chi^')||_2^2 
        \\&~~~~~~~~~~~~~ \leq A^2 L_H^2 \frac{k(t)^2}{Dt}+ L_H^2 \max_{\chi \in \Theta}||\chi-\chi^'||_2^2 \epsilon
        \\&~~~~~~~~~~~~~ \leq 2A^2L_H^2\max_{\theta \in \Theta}||\chi-\chi^'||_2^2 \frac{log\frac{e|\Theta|\pi^2t^2\sqrt{Dt}}{6\delta(e-1)\sqrt{\pi_0(\chi')}}}{Dt}
        \\&~~~~~~~~~~~~~ = O\left(\frac{log Dt+log\frac{1}{}\delta}{Dt} \right)
    \end{aligned}
\end{equation}
Let $\mathscr{E}_t$ denote the event that the above inequality holds, and $\mathscr{E}_t^c$ denote that the above inequality does not hold. 
\begin{equation}
    \mathbb{P}(\mathscr{E}_t^c) \leq \frac{6\delta}{\pi^2t^2}
\end{equation}
Therefore,
\begin{equation}
    \begin{aligned}
        &\mathbb{P}(\cap_{t=1}^\infty \mathscr{E}_t)
        \\&~~~~~~~~~~~~~ = 1 - \mathbb{P}(\cup_{t=1}^\infty \mathscr{E}_t^c)
        \\&~~~~~~~~~~~~~ \geq 1 - \sum_{t=1}^\infty \mathbb{P}(\mathscr{E}_t^c)) ~~~~~~~\text{(Union bound)}
        \\&~~~~~~~~~~~~~ \geq 1 - \sum_{t=1}^\infty \frac{6\delta}{\pi^2t^2}
        \\&~~~~~~~~~~~~~ = 1 - \delta
    \end{aligned}
\end{equation}

\subsection{Proof of Lemma \ref{lemma5}}
From Assumption \ref{Assump:Variance-gradient}, we have
\begin{equation}
\begin{aligned}
\mathbb{E} ||g_k^t- \bar{g}_k^t||^2 &= \mathbb{E} \Big\Vert \frac{1}{N} \sum_{n\in\mathcal{N}} \left(\nabla F(\boldsymbol{w}_{n,k}^t,\chi_{n,k}^t) - \nabla F(\boldsymbol{w}_{n,k}^t) \right)\Big\Vert^2 
\\&=\frac{1}{N^2}   \sum_{n\in\mathcal{N}} \mathbb{E} \Big\Vert\left(\nabla F(\boldsymbol{w}_{n,k}^t,\chi_{n,k}^t) - \nabla F(\boldsymbol{w}_{n,k}^t) \right)\Big\Vert^2
\\&\leq \frac{\sigma_g^2}{N^2}.
\end{aligned}
\end{equation}
\subsection{Proof of Lemma \ref{lemma6}}
We know that in every global communication round, each client performs $T$ rounds of local SGDs where there always exits $t' \leq t$ such that $t-t'\leq T$ and $\boldsymbol{w}_{n,k}^{t'} = \bar{\boldsymbol{w}}_k^{t'}$, $\forall n \in \mathcal{N}$. By using the fact that $\mathbb{E}||X-\mathbb{E}X||^2 = ||X||^2 - ||\mathbb{E}X||^2$ and $\bar{\boldsymbol{w}}_k^t =\mathbb{E}\boldsymbol{w}_{n,k}^t$, we have:
\begin{equation} 
\begin{aligned}
&\frac{1}{N}\sum_{n\in\mathcal{N}} \mathbb{E} \Big\Vert\bar{\boldsymbol{w}}_k^t-\boldsymbol{w}_{n,k}^t\Big\Vert^2 
\\&~~~~~~~~ = \frac{1}{N}\sum_{n\in\mathcal{N}} \mathbb{E} \Big\Vert\boldsymbol{w}_{n,k}^t - \bar{\boldsymbol{w}}_k^t\Big\Vert^2 
\\&~~~~~~~~=\frac{1}{N}\sum_{n\in\mathcal{N}}\mathbb{E} \Big\Vert(\boldsymbol{w}_{n,k}^t - \bar{\boldsymbol{w}}_k^{t'}) - (\bar{\boldsymbol{w}}_k^t-\bar{\boldsymbol{w}}_k^{t'}) \Big\Vert^2 
\\&~~~~~~~~\leq  \frac{1}{N}\sum_{n\in\mathcal{N}} \mathbb{E} \Big\Vert\boldsymbol{w}_{n,k}^t - \bar{\boldsymbol{w}}_k^{t'}\Big\Vert^2  
\\&~~~~~~~~\leq   \frac{1}{N}\sum_{n\in\mathcal{N}}\mathbb{E}\Big\Vert\left(\sum_{t=t'}^{t-1} (\boldsymbol{w}_{n,k}^t-\bar{\boldsymbol{w}}_k^{t'}) \right) \Big\Vert^2  
\\&~~~~~~~~=   \frac{1}{N}\sum_{n\in\mathcal{N}}\mathbb{E}\Big\Vert\left(\sum_{t=t'}^{t-1} \eta_k \nabla F(\boldsymbol{w}_{n,k}^t,\chi_{n,k}^t) \right) \Big\Vert^2 
\\&~~~~~~~~\leq \frac{1}{N}\sum_{n\in\mathcal{N}}\mathbb{E} \Big\Vert\left(\sum_{t=1}^{t-t'} \eta_k \nabla F(\boldsymbol{w}_{n,k}^t,\chi_{n,k}^t) \right) \Big\Vert^2  ,
\end{aligned}
\end{equation}
where the last inequality holds since the learning rate $\eta_k$ is decreasing. Using the fact that $||\sum_{t=1}^{U} z^t||^2 \leq U\sum_{t=1}^{U} ||z^t||^2$, $t-t' \leq T$ and assume that $\eta_k^{t'} \leq 2\eta_k$ and $||\nabla F(\boldsymbol{w}_{n,k}^t,\chi_{n,k}^t)||^2 \leq B^2$ for positive constant $B$, we have 
\begin{equation} 
\begin{aligned}
&  \frac{1}{N}\sum_{n\in\mathcal{N}} \mathbb{E} \Big\Vert\bar{\boldsymbol{w}}_k^t-\boldsymbol{w}_{n,k}^t\Big\Vert^2  
\\&\leq  \frac{1}{N}\sum_{n\in\mathcal{N}}\left( \mathbb{E} \sum_{t=1}^{t-t'} \eta_k^2(t-t')  \Big\Vert \nabla F(\boldsymbol{w}_{n,k}^t,\chi_{n,k}^t)\Big\Vert^2 \right) 
\\&\leq  \frac{1}{N}\sum_{n\in\mathcal{N}}\left( \mathbb{E} \sum_{t=1}^{t-t'} \eta_k^2T \Big\Vert\nabla F(\boldsymbol{w}_{n,k}^t,\chi_{n,k}^t)\Big\Vert^2 \right) 
\\&\leq  \frac{1}{N}\sum_{n\in\mathcal{N}}\left({(\eta_k^{t'})}^2T  \sum_{t=1}^{t-t'}  B^2 \right) 
\\&\leq  \frac{1}{N}\sum_{n\in\mathcal{N}} {(\eta_k^{t'})}^2T B^2 \leq  4\eta_kT B^2.
\end{aligned}
\end{equation}

\subsection{Proof of Theorem \ref{theorem2}}
The local SGD update at client $n$ is followed as:
\begin{equation}
\begin{aligned}
\bar{\boldsymbol{w}}_{k}^{t+1} = &\boldsymbol{w}_{k}^{t} - \eta_k \nabla F(\boldsymbol{w}_{k}^t,\chi_{k}^{'t}) + \eta_k [\mathbb{E}_{\pi_1} \nabla F(\boldsymbol{w}_{k}^t,\chi_{k}^{t})-\\&\nabla F(\boldsymbol{w}_{k}^t,\bar\chi_{k}^{'t})] - \eta_k [\nabla f(\boldsymbol{w}_{k}^t,\xi_{k}^{t})-\mathbb{E}_{\pi_1} \nabla F(\boldsymbol{w}_{k}^t,\bar\chi_{k}^t)],
\end{aligned}
\end{equation}
where $F$ is a local loss function, $\eta >0$ is the local learning rate, and $\chi$ is a sample uniformly chosen from the local dataset. Now if we will consider $\mathbb{E}_{\pi_1} \nabla F(\boldsymbol{w}_{k}^t,\chi_{k}^{t})-\nabla F(\boldsymbol{w}_{k}^t,\bar\chi_{k}^{'t})$ as bias $B_k$ and $\nabla f(\boldsymbol{w}_{k}^t,\xi_{k}^{t})-\mathbb{E}_{\pi_1} \nabla F(\boldsymbol{w}_{k}^t,\bar\chi_{k}^t)$ as noise $N_k$, the equation becomes:

\begin{equation}
    \bar{\boldsymbol{w}}_{k}^{t+1} = \boldsymbol{w}_{k}^{t} - \eta_k \nabla F(\boldsymbol{w}_{k}^t,\chi_{k}^{'t}) + \eta_k B_k - \eta_k N_k,
\end{equation}

By Lemma \ref{lemma3}, we know
\begin{equation}
    \mathbb{E}[||B_t||_2^2] \leq C_1 \frac{log Dt+log\frac{1}{}\delta}{Dt}
\end{equation}
By Assumption \ref{assump:4}, we have
\begin{equation}
    \mathbb{E}[||N_t||_2^2] \leq \sigma^2
\end{equation}
By proof of Lemma \ref{lemma2}, we know that
\begin{equation}
    \begin{aligned}
        &\mathbb{E} [F(\boldsymbol{w}_{k}^{t+1},\chi_{k}^{'t}) - F(\boldsymbol{w}_{k}^t,\chi_{k}^{'t})] \\&\leq -\frac{\eta_k}{2}||\nabla F(\boldsymbol{w}_{k}^t,\chi_{k}^{'t})||_2^2 + \frac{\eta_k}{2}C_1\frac{log Dt+log\frac{1}{\delta}}{Dt}+\frac{\eta_k^2}{2}L_h\sigma^2
    \end{aligned}
\end{equation}
and after multiplying by 2,
\begin{equation}
    \begin{aligned}
        &2\mathbb{E} [F(\boldsymbol{w}_{k}^{t+1},\chi_{k}^{'t}) - F(\boldsymbol{w}_{k}^t,\chi_{k}^{'t})] \\&\leq -\eta_k||\nabla F(\boldsymbol{w}_{k}^t,\chi_{k}^{'t})||_2^2 + \eta_k C_1\frac{log Dt+log\frac{1}{\delta}}{Dt}+\eta_k^2 L_h\sigma^2
    \end{aligned}
\end{equation}
after rearranging,
\begin{equation}
    \begin{aligned}
        &\eta_k||\nabla F(\boldsymbol{w}_{k}^t,\chi_{k}^{'t})||_2^2 \\&\leq -2\mathbb{E} [F(\boldsymbol{w}_{k}^{t+1},\chi_{k}^{'t}) - F(\boldsymbol{w}_{k}^t,\chi_{k}^{'t})] + \eta_k C_1\frac{log Dt+log\frac{1}{\delta}}{Dt}+\\&~~~~\eta_k^2 L_h\sigma^2
    \end{aligned}
\end{equation}
noting that $F(\boldsymbol{w}_{k}^t,\chi_{k}^{'t}) \leq \min_{x in \mathscr{X}} F(\boldsymbol{w},\chi_{k}^{'t})$
\begin{equation}
    \begin{aligned}
        &eta_k||\nabla F(\boldsymbol{w}_{k}^t,\chi_{k}^{'t})||_2^2 \\&\leq 2 (F(\boldsymbol{w}_1,\chi_{k}^{'t})- \min_{x \in \mathscr{X}} F(\boldsymbol{w},\chi_{k}^{'t})) + \eta_k C_1\frac{log Dt+log\frac{1}{\delta}}{Dt}+\\&~~~~\eta_k^2 L_h\sigma^2
    \end{aligned}
\end{equation}
summing over k from 1 to $\mathcal{K}$,
\begin{equation}
    \begin{aligned}
        &\sum_{k=1}^{\mathcal{K}} \eta_k||\nabla F(\boldsymbol{w}_{k}^t,\chi_{k}^{'t})||_2^2 \\&\leq 2 (F(\boldsymbol{w}_1,\chi_{k}^{'t})- \min_{x in \mathscr{X}} F(\boldsymbol{w},\chi_{k}^{'t})) + C_1 \sum_{k=1}^{\mathcal{K}} \eta_k\frac{log Dt+log\frac{1}{\delta}}{Dt}+ \\&~~~~\sigma^2 \sum_{k=1}^{\mathcal{K}} \eta_k^2
    \end{aligned}
\end{equation}
Dividing both sides by $\sum_{k=1}^{\mathcal{K}} \eta_k$,
\begin{equation}
    \begin{aligned}
        &\frac{1}{\sum_{k=1}^{\mathcal{K}} \eta_k} \sum_{k=1}^{\mathcal{K}} \eta_k||\nabla F(\boldsymbol{w}_{k}^t,\chi_{k}^{'t})||_2^2 \\&\leq \frac{1}{\sum_{k=1}^{\mathcal{K}} \eta_k} \bigg[2 (F(\boldsymbol{w}_1,\chi_{k}^{'t})- \min_{x in \mathscr{X}} F(\boldsymbol{w},\chi_{k}^{'t})) + \\&~~~~C_1 \sum_{k=1}^{\mathcal{K}} \eta_k\frac{log Dt+log\frac{1}{\delta}}{Dt} + L_h\sigma^2 \sum_{k=1}^{\mathcal{K}} \eta_k^2 \bigg]
    \end{aligned}
\end{equation}
noting that $\frac{1}{\sum_{k=1}^{\mathcal{K}} \eta_k} \sum_{k=1}^{\mathcal{K}} \eta_k||\nabla F(\boldsymbol{w}_{k}^t,\chi_{k}^{'t})||_2^2 = \mathbb{E}||\nabla F(z_T,\chi_{k}^{'t})||_2^2$,
\begin{equation}
    \begin{aligned}
        &\mathbb{E}||\nabla F(z_T,\chi_{k}^{'t})||_2^2 \\&\leq \frac{1}{\sum_{k=1}^{\mathcal{K}} \eta_k} \bigg[2 (F(\boldsymbol{w}_1,\chi_{k}^{'t})- \min_{x in \mathscr{X}} F(\boldsymbol{w},\chi_{k}^{'t})) + \\&~~~~C_1 \sum_{k=1}^{\mathcal{K}} \eta_k\frac{log Dt+log\frac{1}{\delta}}{Dt} + L_h\sigma^2 \sum_{k=1}^{\mathcal{K}} \eta_k^2 \bigg]
    \end{aligned}
\end{equation}

\begin{enumerate}
        \item[(i)] \setstretch{2} if the step size ($\eta$) satisfies $\eta_k = \frac{a}{\sqrt{K}}, \forall k \leq \mathcal{K}$, for some constant $a < \frac{\sqrt{\mathcal{K}}}{L_h}$.

        Note that $\sum_{k=1}^{\mathcal{K}} \frac{1}{t} \leq log \mathcal{K}+1$ and $\sum_{k=1}^{\mathcal{K}} \frac{log k}{k} \leq log(log \mathcal{K}+1)$. Then
        \begin{equation} \label{eq:delf1}
        \begin{aligned}
        &\Theta_1 \triangleq \mathbb{E}[||\nabla F(z_{\mathcal{K}},\chi_{k}^{'t})||_2^2]
        \\& \leq \frac{2(F(\boldsymbol{w}_1,\chi_{k}^{'t})- \min_{x\in \mathscr{X}}F(\boldsymbol{w},\chi_{k}^{'t}))}{a\sqrt{\mathcal{K}}} + \\&~~~~\frac{C_1(log D-log \delta)(log \mathcal{K}+1)}{L_h D\mathcal{K}}+\frac{C_1 log \mathcal{K}(log \mathcal{K}+1)}{L_hD\mathcal{K}} 
        \\&= \frac{2(F(\boldsymbol{w}_1,\chi_{k}^{'t})- \min_{x\in \mathscr{X}}F(\boldsymbol{w},\chi_{k}^{'t}))}{a\sqrt{\mathcal{K}}} + \frac{C_1(log D-log \delta)}{L_h D\mathcal{K}}+ \\&~~~~\frac{C_1(log D-log \delta)log \mathcal{K}}{L_h D\mathcal{K}} +\frac{C_1 log^2 \mathcal{K}}{L_hD\mathcal{K}}  + \frac{L_ha\sigma^2}{\sqrt{\mathcal{K}}} 
        \end{aligned}
        \end{equation}

        \item[(ii)] if the step size ($\eta$) satisfies $\eta_k = \frac{a}{k}, \forall k \leq \mathcal{K}$, for some constant $a < \frac{1}{L_h}$. Let $M_{\mathcal{K}} = \sum_{k=1}^{\mathcal{K}} \frac{1}{k}$. Note that
        \begin{equation}
            \sum_{k=1}^{\mathcal{K}} \frac{log k}{k^2} < \sum_{k=1}^{\infty} \frac{log k}{k^2} = \frac{\pi^2}{6}(12ln A-\gamma-ln 2\pi) < 1
        \end{equation}
        where the Glaisher-Kinkelin constant $A\approx 1.28$ and the Euler-Mascheroni constant $\gamma \approx 0.58$. Then we have
        \begin{equation} \label{eq:delf2}
        \begin{aligned}
        &\Theta_2 \triangleq \mathbb{E}[||\nabla F(z_{\mathcal{K}},\chi_{k}^{'t})||_2^2]
        \\& \leq \frac{2(F(\boldsymbol{w}_1,\chi_{k}^{'t})- \min_{x\in \mathscr{X}}F(\boldsymbol{w},\chi_{k}^{'t}))}{aM_{\mathcal{K}}} + \\&~~~~\frac{C_1}{M_{\mathcal{K}}} \sum_{k=1}^{\mathcal{K}}\frac{log Dk+ log \frac{1}{\delta}}{Dk^2}+\sum_{k=1}^{\mathcal{K}} \frac{L_ha\sigma^2}{M_{\mathcal{K}}t^2}
        \\& \leq \Bigg[ \frac{2(F(\boldsymbol{w}_1,\chi_{k}^{'t})- \min_{x\in \mathscr{X}}F(\boldsymbol{w},\chi_{k}^{'t}))}{a}+\\&~~~~\frac{6C_1+\pi^2C_1(log D- log \delta)+\frac{\pi^2L_ha\sigma^2}{6}}{6D} \Bigg] \frac{1}{log \mathcal{K}}
        \end{aligned}
        \end{equation}

        \item[(iii)] if the step size ($\eta$) satisfies $\eta_k = \frac{a}{\sqrt{k}}, \forall k \leq \mathcal{K}$, for some constant $a < \frac{1}{L_h}$. Let $Q_t = \sum_{k=1}^{\mathcal{K}} \frac{1}{\sqrt{k}}$. Note that $\sum_{k=1}^{\infty} \frac{1}{k\sqrt{k}} = \zeta(1.5) \approx 2.61 \leq 3$, $\sum_{t=1}^{\infty} \frac{log k}{k\sqrt{k}} < 4, \sum_{k=1}^{\mathcal{K}} \frac{1}{\sqrt{k}}\geq\sqrt{\mathcal{K}}$, where $\zeta(.)$ is the Riemann’s zeta function. Then we have

        \begin{equation} \label{eq:delf3}
        \begin{aligned}
        &\Theta_3 \triangleq \mathbb{E}[||\nabla F(z_{\mathcal{K}},\chi_{k}^{'t})||_2^2]
        \\& \leq \frac{2(F(\boldsymbol{w}_1,\chi_{k}^{'t})- \min_{x\in \mathscr{X}}F(\boldsymbol{w},\chi_{k}^{'t}))}{aQ_{\mathcal{K}}} + \\&~~~~\frac{C_1(log D-log \delta)}{DQ_{\mathcal{K}}} \sum_{k=1}^{\mathcal{K}} \frac{1}{k\sqrt{k}} + \frac{C_1}{DQ_{\mathcal{K}}}\sum_{k=1}^{\mathcal{K}} \frac{log k}{k\sqrt{k}} + \\&~~~~\frac{L_ha\sigma^2}{Q_{\mathcal{K}}} \sum_{k=1}^{\mathcal{K}} \frac{1}{k}
        \\& \leq [\frac{2(F(\boldsymbol{w}_1,\chi_{k}^{'t})- \min_{x\in \mathscr{X}}F(\boldsymbol{w},\chi_{k}^{'t}))}{a\sqrt{\mathcal{K}}} + \\&~~~~\frac{3C_1(log D - log \delta) +4C_1}{D\sqrt{\mathcal{K}}}+\frac{L_ha\sigma^2}{\sqrt{\mathcal{K}}}]+\frac{L_ha\sigma^2log \mathcal{K}}{\sqrt{\mathcal{K}}}
        \end{aligned}
        \end{equation}
\end{enumerate}
\subsection{Proof of Theorem \ref{theorem_final}} 
From the SGD update rule $\bar{\boldsymbol{w}}_k^{t+1}  = \bar{\boldsymbol{w}}_k^t - \eta_kg_k^t + \bar{\textbf{v}}_k^t$ and $||a+b||^2 \leq 2||a||^2 +2||b||^2$ for two real valued vectors $a$ and $b$, we have
\begin{equation}\label{eq:main1}
\begin{aligned}
 ||\bar{\boldsymbol{w}}_k^{t+1} - \boldsymbol{w}^*||^2 &= ||\bar{\boldsymbol{w}}_k^t - \eta_kg_k^t + \bar{\textbf{v}}_k^t - \boldsymbol{w}^*||^2  \\&\leq \underbrace{||\bar{\boldsymbol{w}}_k^t - \eta_kg_k^t  - \boldsymbol{w}^*||^2}_{(A)} + ||\bar{\textbf{v}}_k^t||^2 
\end{aligned}
\end{equation}
We now focus on the bounding term $(A)$ in~\ref{eq:main1}. We have
\begin{equation} \label{equa:termA}
\begin{aligned}
&||\bar{\boldsymbol{w}}_k^t - \eta_kg_k^t  - \boldsymbol{w}^*||^2 = ||\bar{\boldsymbol{w}}_k^t - \eta_kg_k^t - \boldsymbol{w}^* -\eta_k\bar{g}_k^t +\eta_k\bar{g}_k^t||^2 
\\&=||(\bar{\boldsymbol{w}}_k^t - \boldsymbol{w}^* -\eta_k\bar{g}_k^t||^2 + 2\eta_k\langle \bar{\boldsymbol{w}}_k^t - \boldsymbol{w}^* -\eta_k\bar{g}_k^t,\bar{g}_k^t -g_k^t\rangle + \\&~~~~\eta_k^2||g_k^t - \bar{g}_k^t||^2  
\\&=\underbrace{||(\bar{\boldsymbol{w}}_k^t - \boldsymbol{w}^* -\eta_k\bar{g}_k^t||^2}_{(B)}  + \eta_k^2||g_k^t - \bar{g}_k^t||^2,
\end{aligned}
\end{equation}
where $\langle \bar{\boldsymbol{w}}_k^t - \boldsymbol{w}^* -\eta_k\bar{g}_k^t,\bar{g}_k^t -g_k^t\rangle =0$. \\We now focus on bounding term $(B)$. We have 
\begin{equation} \label{equa:boundB1}
\begin{aligned}
&||(\bar{\boldsymbol{w}}_k^t - \boldsymbol{w}^* -\eta_k\bar{g}_k^t||^2 
\\&= ||\bar{\boldsymbol{w}}_k^t - \boldsymbol{w}^*||^2 + \eta_k^2||\bar{g}_k^t||^2 -2\eta_k \frac{1}{N} \sum_{n\in\mathcal{N}} \langle \bar{\boldsymbol{w}}_k^t - \boldsymbol{w}^*,\nabla F(\boldsymbol{w}_{n,k}^t)    \rangle
\\& \leq ||\bar{\boldsymbol{w}}_k^t - \boldsymbol{w}^*||^2 + \eta_k^2\frac{1}{N} \sum_{n\in\mathcal{N}}||\nabla F(\boldsymbol{w}_{n,k}^t)||^2 -\\&~~~~2\eta_k \frac{1}{N} \sum_{n\in\mathcal{N}} \langle \bar{\boldsymbol{w}}_k^t - \boldsymbol{w}_{n,k}^t +\boldsymbol{w}_{n,k}^t - \boldsymbol{w}^*,\nabla F(\boldsymbol{w}_{n,k}^t)  \rangle
\\& \leq ||\bar{\boldsymbol{w}}_k^t - \boldsymbol{w}^*||^2 + 2\eta_k^2\frac{L}{N} \sum_{n\in\mathcal{N}} (F(\boldsymbol{w}_{n,k}^t) - F^*) -\\&~~~~2\eta_k \frac{1}{N} \sum_{n\in\mathcal{N}} \langle \bar{\boldsymbol{w}}_k^t - \boldsymbol{w}_{n,k}^t,\nabla F(\boldsymbol{w}_{n,k}^t) \rangle
\\&~~~~-2\eta_k \frac{1}{N} \sum_{n\in\mathcal{N}} \langle \boldsymbol{w}_{n,k}^t - \boldsymbol{w}^*,\nabla F(\boldsymbol{w}_{n,k}^t) \rangle,
\end{aligned}
\end{equation}
where in the first inequality we applied \[||\sum_{n\in\mathcal{N}} z_n||^2 \leq N\sum_{n\in\mathcal{N}} ||z_n||^2\], and in the second inequality we applied  L-smoothness  \[||\nabla F(\boldsymbol{w}_{n,k}^t)||^2 \leq 2L (F(\boldsymbol{w}_{n,k}^t) - F^*)\]. For the third term in~\ref{equa:boundB1}, by using the Cauchy–Schwarz inequality and arithmetic and geometric means (AM-GM) inequality: \[2\langle a,b \rangle \leq \frac{1}{\varepsilon}||a||^2 +\varepsilon||b||^2\] for $\varepsilon>0$, we have 

\begin{equation}
\begin{aligned}
 &-2 \langle \bar{\boldsymbol{w}}_k^t - \boldsymbol{w}_{n,k}^t,\nabla F(\boldsymbol{w}_{n,k}^t) \rangle 
\\&~~~~~~~= 2 \langle \boldsymbol{w}_{n,k}^t - \bar{\boldsymbol{w}}_k^t,\nabla F(\boldsymbol{w}_{n,k}^t) \rangle 
\\&~~~~~~~\leq \frac{1}{\eta_k} ||\boldsymbol{w}_{n,k}^t - \bar{\boldsymbol{w}}_k^t||^2 + \eta_k||\nabla F(\boldsymbol{w}_{n,k}^t)||^2
\\&~~~~~~~ \leq \frac{1}{\eta_k} ||\boldsymbol{w}_{n,k}^t - \bar{\boldsymbol{w}}_k^t||^2 + 2\eta_k L (F(\boldsymbol{w}_{n,k}^t) - F^*).
\end{aligned}
\end{equation}

For the last term in~\ref{equa:boundB1}, by using $\mu$-strong convexity, we have 
\begin{equation}
\begin{aligned}
&- \langle \boldsymbol{w}_{n,k}^t - \boldsymbol{w}^*,\nabla F(\boldsymbol{w}_{n,k}^t) \rangle \\&\leq -(F(\boldsymbol{w}_{n,k}^t) - F^*) -\frac{\mu}{2}||\boldsymbol{w}_{n,k}^t - \boldsymbol{w}^*||^2.
\end{aligned}
\end{equation}

Therefore, \ref{equa:boundB1} can be rewritten as
\begin{equation}
    \begin{aligned} \label{Equa:subB1}
&||(\bar{\boldsymbol{w}}_k^t - \boldsymbol{w}^* -\eta_k\bar{g}_k^t||^2 
\\& \leq ||\bar{\boldsymbol{w}}_k^t - \boldsymbol{w}^*||^2 + 2\eta_k^2\frac{L}{N} \sum_{n\in\mathcal{N}}(F(\boldsymbol{w}_{n,k}^t) - F^*) 
+ \\&~~~~\eta_k\frac{1}{N} \sum_{n\in\mathcal{N}} \left(\frac{1}{\eta_k} ||\boldsymbol{w}_{n,k}^t - \bar{\boldsymbol{w}}_k^t||^2 + 2\eta_k L (F(\boldsymbol{w}_{n,k}^t) - F^*) \right) -
\\&~~~~2\eta_k \frac{1}{N} \sum_{n\in\mathcal{N}}(F(\boldsymbol{w}_{n,k}^t) - F^*) -\mu\eta_k \frac{1}{N} \sum_{n\in\mathcal{N}}\frac{\mu}{2}||\boldsymbol{w}_{n,k}^t - \boldsymbol{w}^*||^2
\\&\leq ||\bar{\boldsymbol{w}}_k^t - \boldsymbol{w}^*||^2 + 2\eta_k(2\eta_kL-1) \frac{1}{N} \sum_{n\in\mathcal{N}}(F(\boldsymbol{w}_{n,k}^t) - F^*) + \\&~~~~\frac{1}{N} \sum_{n\in\mathcal{N}}||\bar{\boldsymbol{w}}_k^t-\boldsymbol{w}_{n,k}^t||^2  -\mu\eta_k \frac{1}{N} \sum_{n\in\mathcal{N}}||\boldsymbol{w}_{n,k}^t - \boldsymbol{w}^*||^2
\\&= (1-\mu\eta_k)||\bar{\boldsymbol{w}}_k^t - \boldsymbol{w}^*||^2 + 2\eta_k(2\eta_kL-1) \frac{1}{N} \\&~~~~\sum_{n\in\mathcal{N}}(F(\boldsymbol{w}_{n,k}^t) - F^*) + \frac{1}{N}\sum_{n\in\mathcal{N}}||\bar{\boldsymbol{w}}_k^t-\boldsymbol{w}_{n,k}^t||^2,
\end{aligned} 
\end{equation}
where we used the fact: $\frac{1}{N} \sum_{n\in\mathcal{N}}||\boldsymbol{w}_{n,k}^t - \boldsymbol{w}^*||^2 = ||\bar{\boldsymbol{w}}_k^t - \boldsymbol{w}^*||^2$. We assume $\eta_k \leq \frac{1}{4L}$, it holds $\eta_kL \leq \frac{1}{4} \Longrightarrow 2\eta_kL -1 \leq -\frac{1}{2}$. Thus
\begin{equation}
\begin{aligned}
&||(\bar{\boldsymbol{w}}_k^t - \boldsymbol{w}^* -\eta_k\bar{g}_k^t||^2 \\&\leq (1-\mu\eta_k)||\bar{\boldsymbol{w}}_k^t - \boldsymbol{w}^*||^2  + \frac{1}{N}\sum_{n\in\mathcal{N}}||\bar{\boldsymbol{w}}_k^t-\boldsymbol{w}_{n,k}^t||^2-\\&~~~~\frac{1}{2} \frac{1}{N} \sum_{n\in\mathcal{N}}(F(\boldsymbol{w}_{n,k}^t) - F^*)
\end{aligned}
\end{equation}
\begin{equation} \label{equa:Cterm}
\begin{aligned}
&||(\bar{\boldsymbol{w}}_k^t - \boldsymbol{w}^* -\eta_k\bar{g}_k^t||^2 \\&\leq (1-\mu\eta_k)||\bar{\boldsymbol{w}}_k^t - \boldsymbol{w}^*||^2  + \frac{1}{N}\sum_{n\in\mathcal{N}}||\bar{\boldsymbol{w}}_k^t-\boldsymbol{w}_{n,k}^t||^2- \\&~~~~\underbrace{\frac{1}{2} \mathbb{E}[||\nabla F(z_{\mathcal{K}},\chi_{k}^{'t})||_2^2]}_{(C)}
\end{aligned}
\end{equation}
We can bound $(C)$ using equation \ref{eq:delf1}, \ref{eq:delf2}, and \ref{eq:delf3} 
\begin{equation}
\mathbb{E}[||\nabla F(z_{\mathcal{K}},\chi_{k}^{'t})||_2^2] = \min (\Theta_1,\Theta_2,\Theta_3)
\end{equation}
where the first inequality results from the convexity of $F_n(.)$, the second inequality is derived from the AM-GM inequality, and the third inequality results from the smoothness of $F_n(.)$. Therefore,~\ref{equa:Cterm}
is further expressed as
\begin{equation} 
\begin{aligned}\label{equa:subsubC-final}
&||(\bar{\boldsymbol{w}}_k^t - \boldsymbol{w}^* -\eta_k\bar{g}_k^t||^2 \\&\leq (1-\mu\eta_k)||\bar{\boldsymbol{w}}_k^t - \boldsymbol{w}^*||^2  + \frac{1}{N}\sum_{n\in\mathcal{N}}||\bar{\boldsymbol{w}}_k^t-\boldsymbol{w}_{n,k}^t||^2 + 
\\&~~~~\frac{1}{4\eta_k}\frac{1}{N} \sum_{n\in\mathcal{N}}||\boldsymbol{w}_{n,k}^t -  \bar{\boldsymbol{w}}_k^t||^2 + \frac{1}{2} \min (\Theta_1,\Theta_2,\Theta_3)
\end{aligned}
\end{equation}

By plugging \ref{equa:subsubC-final} into \ref{eq:main1} and taking expectation we obtain
\begin{equation} 
\begin{aligned}
&\mathbb{E}||\bar{\boldsymbol{w}}_k^{t+1} - \boldsymbol{w}^*||^2  
\\&\leq ||(1-\mu\eta_k)||\bar{\boldsymbol{w}}_k^t - \boldsymbol{w}^*||^2  + \frac{1}{N}\sum_{n\in\mathcal{N}}||\bar{\boldsymbol{w}}_k^t-\boldsymbol{w}_{n,k}^t||^2 + 
\\&~~~~\frac{1}{4\eta_k}\frac{1}{N} \sum_{n\in\mathcal{N}}||\boldsymbol{w}_{n,k}^t -  \bar{\boldsymbol{w}}_k^t||^2 + \frac{1}{2} \min (\Theta_1,\Theta_2,\Theta_3)  + \\&~~~~\eta_k^2||g_k^t - \bar{g}_k^t||^2
\end{aligned}
\end{equation}

From Lemmas \ref{lemma5}, \ref{lemma6}, and \ref{lemma7}, we have
\begin{equation} \label{equa:updaterule_final}
\begin{aligned}
&\mathbb{E}||\bar{\boldsymbol{w}}_k^{t+1} - \boldsymbol{w}^*||^2  
\\&\leq (1-\mu\eta_k)\mathbb{E}||\bar{\boldsymbol{w}}_k^t - \boldsymbol{w}^*||^2  
+ 4 \left(1+\frac{1}{\eta_k}\right) \eta_kT B^2 + \frac{\eta_k^2\sigma_g^2}{N^2}
\end{aligned}
\end{equation}

Let us define $ Y_k^t = \mathbb{E}||\bar{\boldsymbol{w}}_k^t - \boldsymbol{w}^*||^2$ and $\Phi_k = 4 \left(\frac{\eta_k+1}{\eta_k^2} \right)T B^2 + \frac{\sigma_g^2}{N^2}$, from \ref{equa:updaterule_final} we have
\begin{equation}\label{equa:ytransform}
\sum_{t=1}^{T}Y_k^{t+1} \leq \sum_{t=0}^{T-1} (1-\mu\eta_k)Y_k^t + \eta_k^2\Phi_k,
\end{equation}
By $Y_k = \sum_{t=0}^{T-1}Y_k^t $, \ref{equa:ytransform} is rewritten as
\begin{equation}\label{equa:ykform}
Y_k^{t+1} \leq (1-\mu\eta_k)Y_k^t  + \eta_k^2\Phi_k,   
\end{equation}
We define a diminishing stepsize $\eta_k = \frac{4\theta}{k+\omega}$ for some $\theta >\frac{1}{4\mu}$ and $\omega >0$. By defining $m =\max \{\frac{\theta^2\Phi_k}{4\theta\mu-1}, (\omega+1)Y_0\}$, we prove that $Y_k \leq \frac{m}{k+\omega}$ by induction. Due to $4\theta\mu >1$, from \ref{equa:ykform} we have
\begin{equation} \label{equa:_final}
\begin{aligned}
Y_{k+1} &= \left(1-\frac{4\theta\mu}{k+\omega} \right) \frac{m}{k+\omega} +  \frac{16\theta^2}{(k+\omega)^2}\Phi_k
 \\&\leq \frac{k+\omega-1}{(k+\omega)^2}m + \frac{16\theta^2}{(k+\omega)^2}\Phi_k
\\& \leq \frac{k+\omega-1}{(k+\omega)^2}m + \frac{16\theta^2}{(k+\omega)^2}\Phi_k
- \frac{4\theta\mu-1}{(k+\omega)^2} 
 \\&\leq \frac{k+\omega-1}{(k+\omega)^2}m - \frac{4\theta\mu-1}{(k+\omega)^2}
 \\& \leq \frac{k+\omega-4\theta\mu}{(k+\omega)^2}m  \\&\leq \frac{k+\omega-4\theta\mu}{(k+\omega)^2 - (4\theta\mu)^2}m 
 \\& \frac{1}{k+\omega+4\theta\mu}m \\&\leq \frac{1}{k+\omega+1}m
\end{aligned} 
\end{equation}
We choose $\theta = \frac{4}{\mu}$ and $\omega = \frac{L}{\mu}$ , it follows that
\begin{equation}
\begin{aligned}
    m &=\max \{\frac{\theta^2\Phi_k}{4\theta\mu-1}, (\omega+1)Y_0\} \\&\leq \frac{\theta^2\Phi_k}{4\theta\mu-1} + (\omega+1)Y_0 \\&= \frac{16\Phi_k}{15\mu^2} + \left( \frac{L}{\mu}+1\right)Y_0
\end{aligned}
\end{equation}
By using the $L$-smoothness of $F(.)$, we have
\begin{equation} \label{equa:final_convergenceIID}
\begin{aligned}
    &\mathbb{E}\left[F(\bar{\boldsymbol{w}}_k)\right] -F^* \\&\leq \frac{L}{2}Y_k\leq \frac{L}{2}\frac{m}{(k+\omega)} \\&\leq \frac{L}{2(k+L/\mu)}\Bigg[\frac{16\Phi_k}{15\mu^2} + \left( \frac{L}{\mu}+1\right) \mathbb{E}||\boldsymbol{w}_0 - \boldsymbol{w}^*||^2 \Bigg]
\end{aligned}
\end{equation}
Finally, by applying \ref{equa:final_convergenceIID} recursively, the convergence bound of our approach after $K$ global communication rounds can be given as 
\begin{equation}
\begin{aligned}
    &\mathbb{E}\left[F(\boldsymbol{w}_K)\right] -F^* \\&\leq \frac{L}{2(K+L/\mu)}\left[\frac{16\Phi_K}{15\mu^2} + \left( \frac{L}{\mu}+1\right) \mathbb{E}||\boldsymbol{w}_0 - \boldsymbol{w}^*||^2 \right],
\end{aligned}
\end{equation}
which completes the proof.

{\color{black}
\noindent\textbf{Interpretation of the bound:}
In terms of the step size, stochastic gradient variance, and data-heterogeneity factors, the derived upper limit describes the predicted optimality gap of the global model following $T$ communication rounds, i.e., $\mathbb{E}[F(w_T)] - F(w^\star)$ (or an analogous global loss gap). The bound lowers with $T$ (often showing a $O(1/T)$ dependency) under conventional smoothness and bounded-variance assumptions, indicating better global performance as the number of rounds rises. We emphasize that the constraint is not claimed to be tight; rather, it acts as a convergence guarantee for Meta-BayFL under probabilistic personalized updates.

Even though Theorem~1 provides convergence guarantees for different learning rate schedules, practical observations show that the meta-learned adaptive schedule is used in real life because it consistently produces faster convergence and stronger stability in experimental situations that are not IID.}

\section{Simulations and Performance Evaluation}
\subsection{Environment Settings}
For our \textit{Meta-BayFL} approach, we only consider non-IID data. Clients have different data sizes, data points, number of features, and batch sizes. The total number of clients is five, and we have only considered the non-IID data. For our simulation, we have used three datasets: i) CIFAR-10 \cite{krizhevsky2010convolutional}, ii) CIFAR-100 \cite{krizhevsky2009learning}, and iii) Tiny-ImageNet \cite{le2015tiny} datasets to calculate the performance of our proposed method. 
\begin{table*}
\footnotesize
    \centering
    \begin{tabular}{|c|c|c|c|c|c|c|c|}
        \hline
        \textbf{Non-i.i.d. Type} & $\mathcal{U}$ & $|\mathcal{U}|$ & \textbf{ConvNet} & \textbf{ResNet20} & \textbf{ResNet32} & \textbf{ResNet44} & \textbf{ResNet56} \\
        \hline
        \multirow{3}{*}{Step} 
        & CIFAR-10 & 10K & 65.5$\pm$0.52 & 67.5$\pm$0.52 & 62.7$\pm$0.47 & 55.5$\pm$0.42 & 57.0$\pm$0.55 \\
        & CIFAR-100 & 50K & 65.4$\pm$0.48 & 68.2$\pm$0.48 & 62.3$\pm$0.33 & 55.1$\pm$0.41 & 51.0$\pm$0.59 \\
        & Tiny-ImageNet & 100K & 65.5$\pm$0.60 & 67.1$\pm$0.61 & 62.3$\pm$0.53 & 54.5$\pm$0.41 & 59.0$\pm$0.42 \\
        \hline
        \multirow{3}{*}{Dirichlet} 
        & CIFAR-10 & 10K & 63.9$\pm$0.55 & 68.2$\pm$0.46 & 67.7$\pm$0.55 & 57.0$\pm$0.48 & 57.0$\pm$0.42 \\
        & CIFAR-100 & 50K & 63.5$\pm$0.51 & 68.6$\pm$0.53 & 66.5$\pm$0.41 & 62.0$\pm$0.27 & 56.9$\pm$0.47 \\
        & Tiny-ImageNet & 100K & 64.0$\pm$0.55 & 68.2$\pm$0.52 & 56.0$\pm$0.62 & 61.6$\pm$0.76 & 57.3$\pm$0.42 \\
        \hline
    \end{tabular}
    \caption{\footnotesize \textcolor{black}{\textit{Meta-BayFL} on non-i.i.d CIFAR-10 with different unlabeled data $U$. rows correspond to non-IID types (Step/Dirichlet) and choices of unlabeled data source and size; entries report accuracy (mean $\pm$ std) for each backbone. Out-of-domain unlabeled data (e.g., CIFAR-100/Tiny-ImageNet) is sufficient to maintain or improve performance, supporting the practicality of the approach.}}
    \label{tab:unlabeled}
\end{table*}

\begin{table*}[!ht]
\footnotesize
    \centering
    \begin{tabular}{|c|c|cc|cc|cc|}
    \toprule
    \multicolumn{2}{|c|}{\multirow{2}{*}{Data}} & \multicolumn{2}{c|}{\textbf{CIFAR-10}} & \multicolumn{2}{c|}{\textbf{CIFAR-100}} & \multicolumn{2}{c|}{\textbf{Tiny-ImageNet}} \\
    \cmidrule{3-8}
    \multicolumn{2}{|c|}{} & \textbf{FedAVG} & \textbf{BayFL} & \textbf{FedAVG} & \textbf{BayFL} & \textbf{FedAVG} & \textbf{BayFL} \\
    \midrule
    \multirow{6}{*}{$|D|$ = 100\%} & $\epsilon$ = 0.0 & 66.86\% & 67.20\% & 71.25\% & 71.23\% & 73.11\% & 73.59\% \\
    & $\epsilon$ = 0.0001 & 66.71\% & 66.81\% & 71.22\% & 71.20\% & 73.05\% & 73.50\% \\
    & $\epsilon$ = 0.001 & 65.69\% & 66.69\% & 69.50\% & 71.05\% & 72.20\% & 73.44\% \\
    & $\epsilon$ = 0.01 & 63.10\% & 66.52\% & 67.98\% & 70.60\% & 70.66\% & 73.20\% \\
    & $\epsilon$ = 0.1 & 59.23\% & 65.65\% & 64.40\% & 70.00\% & 67.84\% & 72.19\% \\
    \midrule
    \multirow{6}{*}{$|D|$ = 50\%} & $\epsilon$ = 0.0 & 58.28\% & 58.31\% & 59.98\% & 60.03\% & 60.45\% & 60.51\% \\
    & $\epsilon$ = 0.0001 & 56.73\% & 57.14\% & 58.72\% & 59.49\% & 58.66\% & 59.99\% \\
    & $\epsilon$ = 0.001 & 53.47\% & 56.71\% & 56.50\% & 59.33\% & 56.84\% & 59.33\% \\
    & $\epsilon$ = 0.01 & 51.70\% & 55.78\% & 54.67\% & 58.20\% & 55.45\% & 58.70\% \\
    & $\epsilon$ = 0.1 & 50.01\% & 54.65\% & 52.07\% & 57.01\% & 52.77\% & 57.55\% \\
    \midrule
    \multirow{6}{*}{$|D|$ = 25\%} & $\epsilon$ = 0.0 & 49.39\% & 49.78\% & 48.32\% & 49.83\% & 51.22\% & 51.20\% \\
    & $\epsilon$ = 0.0001 & 48.31\% & 49.88\% & 48.44\% & 49.49\% & 50.12\% & 50.92\% \\
    & $\epsilon$ = 0.001 & 47.47\% & 49.01\% & 46.05\% & 49.33\% & 47.78\% & 50.20\% \\
    & $\epsilon$ = 0.01 & 45.70\% & 48.78\% & 43.68\% & 48.19\% & 45.80\% & 49.66\% \\
    & $\epsilon$ = 0.1 & 42.99\% & 48.65\% & 40.70\% & 47.02\% & 43.31\% & 49.19\% \\
    \midrule
    \multirow{6}{*}{$|D|$ = 10\%} & $\epsilon$ = 0.0 & 42.40\% & 45.78\% & 42.32\% & 44.83\% & 41.23\% & 47.02\% \\
    & $\epsilon$ = 0.0001 & 42.18\% & 44.88\% & 42.44\% & 45.49\% & 40.87\% & 46.44\% \\
    & $\epsilon$ = 0.001 & 41.47\% & 44.51\% & 41.50\% & 45.33\% & 40.05\% & 46.04\% \\
    & $\epsilon$ = 0.01 & 40.70\% & 44.38\% & 40.67\% & 43.70\% & 40.01\% & 45.50\% \\
    & $\epsilon$ = 0.1 & 39.91\% & 44.05\% & 38.06\% & 44.00\% & 37.98\% & 44.99\% \\
    \bottomrule
    \end{tabular}  
    \caption{\footnotesize \textcolor{black}{Difference in performance with different data size $|D|$ and noise level $\epsilon$ across various datasets for 5 non-IID Dirichlet clients. Each block fixes $|D|$ and sweeps $\epsilon$; values compare FedAVG vs.\ BayFL on each dataset. The BayFL advantage widens as data becomes smaller and noisier, showing stronger robustness in challenging federated environments.}}
    \label{Tab: smallMnist}
\end{table*}

\begin{figure}
\centering
\includegraphics[width=3 in]{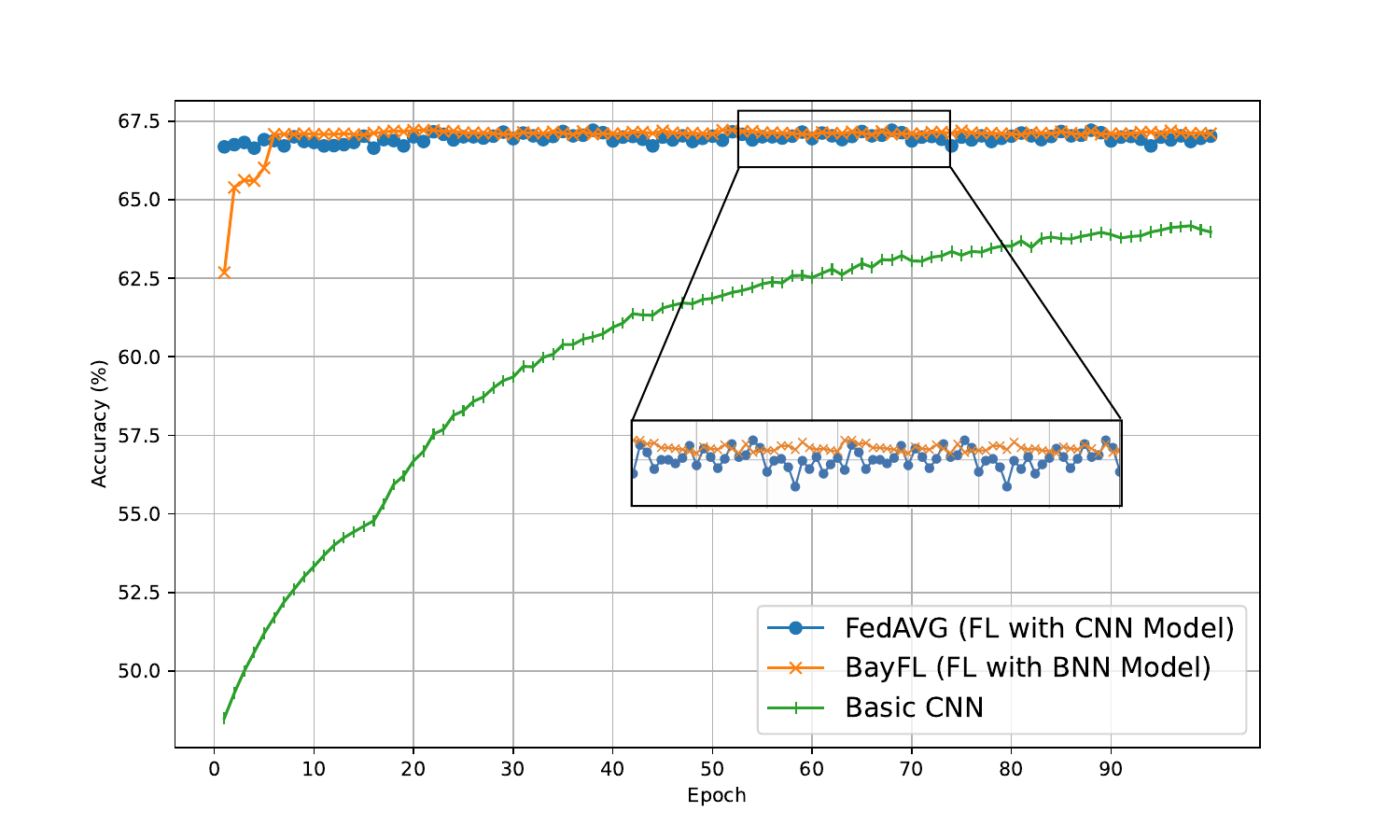}
\caption{\footnotesize \textcolor{black}{Performance comparison on the CIFAR-10 non-IID dataset between basic CNN, FedAVG (FL with CNN), and BayFL (FL with BNN) approaches. A total of 5 clients with various data distributions and batch sizes are used for FL approaches with 10 local epochs in every global round.}}
\vspace{-1mm}
\label{Fig: cvb_mnist}
\end{figure} 

\subsection{Non-IID Data Distribution Setup}
We focus on CIFAR-10, where 10,000 training photos are stored as unlabeled data on the server. The remaining photos are given to ten clients in two non-identifiable scenarios. According to the strategy proposed by Cao et al. \cite{cao2019learning}, in the 'Step' scenario, each client receives 10 images for each of the eight minor classes and 1,960 images for each of the two major classes. In the 'Dirichlet' scenario, suggested in Hsu et al. \cite{hsu2019measuring}, we apply a heterogeneous partitioning method for $N$ clients across $C$ classes. For each class $c \in C$, we create an $N$-dimensional vector $q_c$ using a Dirichlet distribution with parameter 0.1 and distribute the photos to client n based on the fraction of $q_c[n]$. This results in clients having differing total amounts of images.

\subsection{Simulation Results}
\textbf{Comparison with baseline.} First, we compare the performance between basic CNN, FedAVG (basic FL with CNN), and BayFL (FL with BNN) for 5 clients non-IID data in Fig. \ref{Fig: cvb_mnist}. 
The figure shows that both FL (FedAVG and BayFL) significantly outperform CNN. The accuracy results between FedAVG and BayFL are extremely close however, in contrast to FedAVG, BayFL is extremely smooth in the non-IID scenario. Thus, we may say that BNN is more capable than CNN at handling non-IID datasets for the FL settings. However, in the initial epochs, the performance of the BNN approach was worse than that of the CNN approach. It is because in order to begin using Bayesian approaches, one must first create a prior distribution $p(\boldsymbol{w}_{n,k}^{(t)})$ on the weights that represent the degree of uncertainty over their accurate values \cite{shridhar2019comprehensive}. The posterior distributions $q(\boldsymbol{w}_{n,k}^{(t)}|D)$ are still in the early stages of training and have not yet converged to more peaked distributions around the most likely weights. Because of the initial high level of uncertainty, more conservative projections may result from poorly calibrated data, which usually reduces initial accuracy.

\textbf{Effects on unlabeled data.} We investigate the situation in which the unlabeled data comes from a different domain or task. The setup is intended to imitate scenarios in which (a) the server has little knowledge of the client's data and (b) the server is unable to collect unlabeled data that accurately represents the test data. In Table \ref{tab:unlabeled}, we replace the unlabeled data with CIFAR-100 and Tiny-ImageNet. The findings reveal that accuracy remains equivalent to, or even surpasses, that of CIFAR-10, implying that out-of-domain unlabeled data are sufficient for \textit{Meta-BayFL}. 


\textbf{Effects on challenging environment.} However, a distinction is beginning to emerge between the probabilistic approach (BayFL) and the deterministic approach (FedAVG) in Table \ref{Tab: smallMnist} when we employ FedAVG and BayFL on small and noisy different datasets. The results demonstrate that when the dataset size or noise level drops, BayFL consistently surpasses FedAVG, and the performance difference widens. BayFL outperforms FedAVG by 4.14\% on CIFAR-10, 5.94\% on CIFAR-100, and 7.01\% on Tiny-ImageNet when $|D| = 10\%$ and $\epsilon = 0.1$, respectively. Similarly, for $|D| = 25\%$ and $\epsilon = 0.1$, BayFL retains a lead of 5.66\%, 6.32\%, and 5.88\%, respectively. These findings show that BayFL manages uncertainty and limited data more effectively offering it a viable option for federated learning in challenging environments.

\begin{table}
\footnotesize
    \centering
    \begin{tabular}{|p{0.6cm}|c|p{1cm}|p{1cm}|p{1.3cm}|}
    \toprule
    \multicolumn{2}{|c|}{Variables} & \textbf{CIFAR-10} & \textbf{CIFAR-100} & \textbf{Tiny-ImageNet} \\
    
    \midrule
    \multirow{4}{*}{\makecell{$\epsilon$ \\ 0.001}} & BayFL, $lr$ = 0.0001 & 66.69\% & 71.05\% & 73.44\% \\
    & BayFL, $lr$ = 0.001 & 66.90\% & 71.66\% & 73.75\% \\
    & BayFL, $lr$ = 0.01 & 66.02\% & 70.76\% & 73.41\% \\
    & \textbf{\textit{Meta-BayFL}} & \textbf{72.25}\% & \textbf{73.06}\% & \textbf{77.00}\% \\
    \midrule
    \multirow{4}{*}{\makecell{$\epsilon$ \\ 0.01}} & BayFL, $lr$ = 0.0001 & 65.40\% & 70.05\% & 71.34\% \\
    & BayFL, $lr$ = 0.001 & 65.60\% & 70.33\% & 71.30\% \\
    & BayFL, $lr$ = 0.01 & 65.12\% & 70.35\% & 71.25\% \\
    & \textbf{\textit{Meta-BayFL}} &  \textbf{71.67}\% &  \textbf{72.88}\% &  \textbf{76.32}\% \\
    \midrule
    \multirow{4}{*}{\makecell{$\epsilon$ \\ 0.1}} & BayFL, $lr$ = 0.0001 & 62.48\% & 68.80\% & 70.98\% \\
    & BayFL, $lr$ = 0.001 & 62.48\% & 69.20\% & 71.05\% \\
    & BayFL, $lr$ = 0.01 & 62.75\% & 69.02\% & 70.90\% \\
    & \textbf{\textit{Meta-BayFL}} &  \textbf{70.55}\% &  \textbf{71.39}\% &  \textbf{75.81}\% \\
    \bottomrule
    \end{tabular}  
    \caption{\footnotesize \textcolor{black}{Difference in performance with different noise level $\epsilon$ and BayFL with different learning rates $\mathrm{lr}$ across various datasets for 5 non-IID Dirichlet clients. For each $\epsilon$, we compare fixed learning rates against Meta-BayFL. Meta-learning is consistently best and degrades more gracefully under increasing noise.}}
    \label{tab:pfl}
\end{table}

\textbf{Advantage of meta-learning.}
For our meta-learning approach, 
Table \ref{tab:pfl} shows the comparison of results with and without meta-learning for different noise values across all datasets (CIFAR-10, CIFAR-100, and Tiny-ImageNet). \textit{Meta-BayFL} consistently beats BayFL for all datasets and noise levels ($\epsilon$ = 0.001, 0.01, 0.1). As noise grows, accuracy decreases for all models, but \textit{Meta-BayFL} retains a greater accuracy, proving its robustness and resilience to noise. Among BayFL models, a learning rate of 0.001 outperforms 0.0001 and 0.01 in most circumstances. The accuracy reduction is particularly dramatic at $\epsilon = 0.1$, when the basic BayFL models struggle, but \textit{Meta-BayFL} still performs competitively. Overall, \textit{Meta-BayFL} outperforms other approaches, notably in terms of noise management and accuracy stability.

\begin{table}[t]
\centering
\footnotesize
\setlength{\tabcolsep}{2pt}
\begin{tabular}{llcc}
\toprule
\textbf{Environment} & \textbf{Method} & \textbf{Acc (\%)} & \textbf{Time/Round (s)} \\
\midrule
\multirow{3}{*}{Simulated}
& FedAvg / FedProx & 66.85 & 8.60 \\
& Meta-BayFL (MC Dropout) & 67.10 & \textbf{8.72} \\
& \textbf{Meta-BayFL (BNN)} & \textbf{69.48} & 11.00 \\
\midrule
\multirow{3}{*}{Raspberry Pi}
& FedAvg / FedProx & 66.10 & 27.8 \\
& Meta-BayFL (MC Dropout) & 66.40 & \textbf{29.4} \\
& \textbf{Meta-BayFL (BNN)} & \textbf{68.90} & 37.6 \\
\bottomrule
\end{tabular}
\caption{\footnotesize Ablation study comparing uncertainty approximation methods under non-IID CIFAR-10 (Dirichlet $\alpha=0.5$, 5 clients, \textbf{ResNet-56 backbone}, same training budget). Results are reported for both simulated execution and real Raspberry Pi (CPU-only) deployment. Runtime denotes average client-side training time per communication round.}
\label{tab:ablation_uncertainty}
\end{table}

{\color{black}
\textbf{Ablation Study on Uncertainty Approximation.}
We conduct an ablation comparing complete Bayesian neural networks with Monte Carlo dropout under identical non-IID scenarios in Table~\ref{tab:ablation_uncertainty} in order to investigate the trade-off between computational efficiency and uncertainty modeling fidelity. The full BNN-based Meta-BayFL achieves higher accuracy and more stable convergence, justifying its application in extremely heterogeneous and noisy federated situations, while MC dropout shortens runtime. Due to Bayesian parameterization, Meta-BayFL has a little longer runtime than FedAvg/FedProx, but it achieves a notably greater accuracy for strongly non-IID data. The Raspberry Pi results confirm consistent behavior on realistic CPU-only edge hardware by showing the same relative accuracy and runtime ordering as the simulated setting.
}

\begin{table}
\footnotesize
    \centering
    \begin{tabular}{|c|p{1.2cm}|p{1.2cm}|p{1.5cm}|}
    \toprule
    \textbf{Clients} & \textbf{CIFAR-10} & \textbf{CIFAR-100} & \textbf{Tiny-ImageNet} \\
    \midrule
    5   & 72.25\% & 73.06\% & 77.00\% \\
    10  & 73.10\% & 74.25\% & 78.02\% \\
    15  & 73.55\% & 74.80\% & 78.40\% \\
    20  & 74.02\% & 75.10\% & 78.85\% \\
    \bottomrule
    \end{tabular}  
    \caption{\footnotesize \textcolor{black}{Performance of Meta-BayFL under different numbers of non-IID Dirichlet clients with noise level $\epsilon = 0.001$. Each row changes client count while keeping the noise fixed; entries are final test accuracy. Meta-BayFL scales positively with more clients (more diversity), with the expected trade-off of higher system overhead discussed in Section~V.}}
    \label{tab:client_scalability}
\end{table}
\textbf{Scalability with number of clients.} Table~\ref{tab:client_scalability} shows Meta-BayFL's performance with varying client counts (5, 10, 15, and 20) with a fixed noise level ($\epsilon = 0.001$).  We see a noticeable improvement in accuracy as the number of clients grows across all datasets.  This tendency may be ascribed to the increasing data diversity and richer representation that result from more clients contributing to the global model, which enhances generalization performance.  For example, using CIFAR-10, accuracy increases from 72.25\% with 5 clients to 74.02\% with 20 clients. Similarly, CIFAR-100 and Tiny-ImageNet exhibit consistent gains, achieving 75.10\% and 78.85\% accuracy, respectively, with 20 clients.  These findings illustrate Meta-BayFL's capacity to scale to a larger federated population while retaining robust performance in the face of uncertain and diverse data.  However, we observe that increasing the number of clients may result in greater communication costs and synchronization delay, implying a trade-off between accuracy improvements and system overhead.

\begin{table*}[!ht]
\footnotesize
    \centering
    \caption{Comparison with state-of-the-art approaches. }
        \label{Table:final}
    \begin{tabular}{|c|c|c|c|c|c|c|}
    \hline
    \multirow{2}{*}{\textbf{Method}} & \multicolumn{2}{c|}{\textbf{CIFAR-10}} & \multicolumn{2}{c|}{\textbf{CIFAR-100}} & \multicolumn{2}{c|}{\textbf{Tiny-ImageNet}} \\
    \cline{2-7}
    & \textbf{Loss Value} & \textbf{Accuracy} & \textbf{Loss Value} & \textbf{Accuracy} & \textbf{Loss Value} & \textbf{Accuracy} \\
    \hline
    FedAVG \cite{li2019convergence} & 4.35 & 45.50\% & 4.42 & 37.95\% & 4.43 & 36.55\% \\
    BNN \cite{jospin2022hands} & 2.53 & 52.22\% & 2.89 & 48.48\% & 2.75 & 50.18\% \\
    pFedBayes \cite{zhang2022personalized} & 1.94 & 64.27\% & 2.33 & 65.31\% & 2.10 & 63.51\% \\
    pFedBe \cite{yu2025pfedbl} & 1.92 & 64.86\% & 1.99 & 64.03\% & 2.13 & 63.11\% \\
    Fedmask \cite{li2021fedmask} & 1.67 & 66.25\% & 1.88 & 65.50\% & 1.85 & 66.33\% \\
    \textbf{\textit{Meta-BayFL}} & \textbf{1.28} & \textbf{69.48}\% & \textbf{1.33} & \textbf{70.20}\% & \textbf{1.47} & \textbf{73.75}\% \\
    \hline
    \end{tabular}  

\end{table*}



\textbf{Comparison with existing approaches.} Finally, we have compared all our approaches with all existing methods for both noisy and small data in Table \ref{Table:final}. For comparison, we have selected FedAVG \cite{li2019convergence} as the baseline FL method, BNN \cite{jospin2022hands} for the probabilistic approach, FedBE \cite{chen2020fedbe} and FedFomo \cite{zhang2020personalized} for federated probabilistic approach, and Fedmask \cite{li2021fedmask} for PFL with BNN approach to compare with our \textit{Meta-BayFL} approach across all datasets (CIFAR-10, CIFAR-100, and Tiny-ImageNet) in terms of both accuracy and loss value. \textcolor{black}{Despite the fact that both FedMask and Meta-BayFL use Bayesian components, Meta-BayFL performs better due to the synergistic combination of meta-learned personalization and Bayesian uncertainty modeling, which allows for client-adaptive and uncertainty-aware optimization in the presence of heterogeneous data. The meta-learning component allows the model to generalize more effectively among varied federated clients, while the Bayesian framework provides robust uncertainty estimation, mitigating the impact of noise and small sample numbers. \textit{Meta-BayFL} outperforms Fedmask, the second-best performing approach, by 3.23\% in CIFAR-10, 4.70\% in CIFAR-100, and 7.42\% in Tiny-ImageNet, respectively. This suggests that \textit{Meta-BayFL} is a better alternative for federated learning in realistic noisy and small data scenarios.}

{\color{black}
\subsection{Strengths and Limitations.}
The proposed Meta-BayFL framework has a number of noteworthy advantages. First, it successfully tackles client heterogeneity and data uncertainty, two significant issues in actual federated learning systems, by combining Bayesian neural networks with meta-learning-based personalization. Second, a thorough convergence analysis backs up the proposed approach and provides theoretical guarantees on performance and stability throughout communication rounds. Third, even in small, noisy, and non-IID data settings, comprehensive experimental results on CIFAR-10, CIFAR-100, and Tiny-ImageNet show consistent performance improvements over the state-of-the-art federated and personalized learning baselines.

Despite these benefits, our study has a number of limitations that identify areas for further research. One drawback of Meta-BayFL is that, in contrast to deterministic federated learning, the use of Bayesian neural networks results in higher model parameterization and communication overhead. This could restrict direct deployment on IoT devices with extremely limited resources without the need for additional model compression or lightweight uncertainty approximations. Lastly, standard federated learning communicates a single set of deterministic model parameters per communication round, which results in a communication cost. On the other hand, Meta-BayFL increases the per-round communication overhead by about two times by transmitting the Bayesian neural network's mean and variance parameters. Although the load might be decreased by using Bayesian compression and parameter quantization approaches, a promising direction for future research.
}

\section{Conclusion}
In this paper, we propose a novel \textit{Meta-BayFL} methodology to handle uncertain and heterogeneous data based on an innovative design of BNN and pFL. The model increases client diversity by allowing clients with different data and batch sizes to participate in the FL training and create the best local model by fine-tuning their learning rates. We also analyze enhancement in performance on small data by using BNN and on noisy data by using pFL. Finally, we show that \textit{Meta-BayFL} outperforms other state-of-the-art pFL methods in both classification and regression tasks on CIFAR-10, CIFAR-100, and Tiny-ImageNet data respectively over non-IID settings. One major disadvantage of \textit{Meta-BayFL} is that it may lead to high computation on clients due to personalized training. 

{\color{black}
\textbf{Developmental Trends and Open Challenges.}
Current federated learning research focuses on personalized and uncertainty-aware models for diverse edge settings. However, significant hurdles remain in lowering communication overhead, scaling to large and dynamic client populations, and enabling implementation on resource-constrained IoT devices, motivating further research into lightweight uncertainty modeling and compression-aware personalization. Wireless federated learning systems may be vulnerable to communication-layer challenges like jamming in addition to data heterogeneity and unpredictability. Bayesian game-theoretic formulations under partial knowledge have been used in previous works to study jamming mitigation. Such security techniques complement probabilistic personalized federated learning and are an essential factor to consider for realistic deployments.
}


\bibliography{ref.bib}
\bibliographystyle{IEEEtran}
\end{document}